\documentclass[sigconf]{acmart}

\usepackage{amsmath}
\makeatletter
\let\Bbbk\@undefined
\makeatother
\usepackage{amssymb}
\usepackage{dsfont}
\usepackage{booktabs}
\usepackage{subcaption}
\usepackage{multirow}
\usepackage[most]{tcolorbox}
\usepackage{xcolor}
\AtBeginDocument{%
  }

\setcopyright{acmlicensed}
\copyrightyear{2026}
\acmYear{2026}
\acmDOI{XXXXXXX.XXXXXXX}
\acmConference[KDD]{ACM SIGKDD Conference on Knowledge Discovery and Data Mining}{Aug 09--13,
  2026}{Jeju, Korea}
\acmISBN{978-1-4503-XXXX-X/2026/08}

\begin{document}

%%
%% The "title" command has an optional parameter,
%% allowing the author to define a "short title" to be used in page headers.
\title{\textsc{MirrorBench}: A Benchmark to Evaluate Conversational User-Proxy Agents for Human-Likeness}

%%
%% The "author" command and its associated commands are used to define
%% the authors and their affiliations.
%% Of note is the shared affiliation of the first two authors, and the
%% "authornote" and "authornotemark" commands
%% used to denote shared contribution to the research.
\author{Ashutosh Hathidara}
\authornotemark[1]
\affiliation{%
  \institution{SAP Labs}
  \city{Palo Alto}
  \state{California}
  \country{USA}
}
\email{ashutosh.hathidara@sap.com}

\author{Julien Yu}
\affiliation{%
  \institution{SAP Labs}
  \city{Palo Alto}
  \state{California}
  \country{USA}
}

\author{Vaishali Senthil}
\affiliation{%
  \institution{SAP Labs}
  \city{Palo Alto}
  \state{California}
  \country{USA}
}

\author{Sebastian Schreiber}
\affiliation{%
  \institution{SAP Labs}
  \city{Palo Alto}
  \state{California}
  \country{USA}
}

\author{Anil Babu Ankisettipalli}
\affiliation{%
  \institution{SAP Labs}
  \city{Palo Alto}
  \state{California}
  \country{USA}
}

% \author{Valerie B\'eranger}
% \affiliation{%
%   \institution{Inria Paris-Rocquencourt}
%   \city{Rocquencourt}
%   \country{France}
% }

% \author{Aparna Patel}
% \affiliation{%
%  \institution{Rajiv Gandhi University}
%  \city{Doimukh}
%  \state{Arunachal Pradesh}
%  \country{India}}

% \author{Huifen Chan}
% \affiliation{%
%   \institution{Tsinghua University}
%   \city{Haidian Qu}
%   \state{Beijing Shi}
%   \country{China}}

% \author{Charles Palmer}
% \affiliation{%
%   \institution{Palmer Research Laboratories}
%   \city{San Antonio}
%   \state{Texas}
%   \country{USA}}
% \email{cpalmer@prl.com}

% \author{John Smith}
% \affiliation{%
%   \institution{The Th{\o}rv{\"a}ld Group}
%   \city{Hekla}
%   \country{Iceland}}
% \email{jsmith@affiliation.org}

% \author{Julius P. Kumquat}
% \affiliation{%
%   \institution{The Kumquat Consortium}
%   \city{New York}
%   \country{USA}}
% \email{jpkumquat@consortium.net}

%%
%% By default, the full list of authors will be used in the page
%% headers. Often, this list is too long, and will overlap
%% other information printed in the page headers. This command allows
%% the author to define a more concise list
%% of authors' names for this purpose.
% \renewcommand{\shortauthors}{Trovato et al.}

%%
%% The abstract is a short summary of the work to be presented in the
%% article.
\begin{abstract}
Large language models (LLMs) are increasingly used as human simulators, both for evaluating conversational systems and for generating fine-tuning data. However, naive ``act-as-a-user'' prompting often yields verbose, unrealistic utterances, motivating principled evaluation of \emph{user proxy agents}. We present \textsc{MirrorBench}, a reproducible and extensible benchmarking framework that evaluates user proxies solely on their ability to produce human-like user utterances across diverse conversational regimes, explicitly decoupled from downstream task success. \textsc{MirrorBench} combines three lexical-diversity metrics (\textsc{MATTR}, \textsc{Yule’s~$K$}, and \textsc{HD-D}) with three LLM-judge-based metrics (\textsc{GTEval}, \textsc{Pairwise Indistinguishability}, and \textsc{Rubric-and-Reason}), and contextualizes judge scores using Human–Human and Proxy–Proxy calibration controls. Across four public datasets, \textsc{MirrorBench} yields variance-aware comparisons and reveals systematic gaps between user proxies and real human users. The framework is open source\footnote{\url{https://github.com/SAP/mirrorbench}}
 and includes a command-line interface for running and managing user-proxy benchmarking experiments.
 
% Large language models (LLMs) are increasingly used as human simulators, both for evaluating conversational systems and for generating fine-tuning data. However, naive ``act-as-a-user" prompting often yields verbose, unrealistic utterances, underscoring the need for principled evaluation of so-called \emph{user proxy agents}. We present \textsc{MirrorBench}, a reproducible, extensible benchmarking framework that evaluates user proxies solely on their ability to produce human-like user utterances across diverse conversational tasks, explicitly decoupled from downstream task success. \textsc{MirrorBench} combines three lexical-diversity metrics (\textsc{MATTR}, \textsc{Yule’s~$K$}, and \textsc{HD-D}) and three LLM-judge–based metrics (\textsc{GTEval}, \textsc{Pairwise Indistinguishability}, and \textsc{Rubric-and-Reason}), and contextualizes judge scores using Human-Human and Proxy-Proxy calibration controls. Across four open datasets, \textsc{MirrorBench} yields variance-aware results and reveals systematic gaps between user proxies and real human users. The framework is open source\footnote{\url{https://github.com/SAP/mirrorbench}} and includes a command-line interface for running \& managing benchmarking experiments for user-proxy evaluation.
\end{abstract}

%%
%% The code below is generated by the tool at http://dl.acm.org/ccs.cfm.
%% Please copy and paste the code instead of the example below.
%%
\begin{CCSXML}
<ccs2012>
   <concept>
       <concept_id>10010147.10010178.10010219.10010221</concept_id>
       <concept_desc>Computing methodologies~Intelligent agents</concept_desc>
       <concept_significance>500</concept_significance>
       </concept>
   <concept>
       <concept_id>10010147.10010178.10010179.10010181</concept_id>
       <concept_desc>Computing methodologies~Discourse, dialogue and pragmatics</concept_desc>
       <concept_significance>300</concept_significance>
       </concept>
   <concept>
       <concept_id>10010147.10010178.10010179.10010184</concept_id>
       <concept_desc>Computing methodologies~Lexical semantics</concept_desc>
       <concept_significance>500</concept_significance>
       </concept>
   <concept>
       <concept_id>10010147.10010178.10010179.10010182</concept_id>
       <concept_desc>Computing methodologies~Natural language generation</concept_desc>
       <concept_significance>300</concept_significance>
       </concept>
   <concept>
       <concept_id>10010147.10010341.10010370</concept_id>
       <concept_desc>Computing methodologies~Simulation evaluation</concept_desc>
       <concept_significance>300</concept_significance>
       </concept>
 </ccs2012>
\end{CCSXML}

\ccsdesc[500]{Computing methodologies~Intelligent agents}
\ccsdesc[300]{Computing methodologies~Discourse, dialogue and pragmatics}
\ccsdesc[500]{Computing methodologies~Lexical semantics}
\ccsdesc[300]{Computing methodologies~Natural language generation}
\ccsdesc[300]{Computing methodologies~Simulation evaluation}

%%
%% Keywords. The author(s) should pick words that accurately describe
%% the work being presented. Separate the keywords with commas.
\keywords{Large Language Models (LLMs), User-proxy agents, Human-likeness benchmarking, LLM-as-a-judge, Judge sensitivity, Human–LLM correlation, Lexical diversity, MATTR, HD-D, Yule’s K, Pairwise indistinguishability, GTEval, Rubric-And-Reason, ChatbotArena, ClariQ, QULAC, OASST1}
%% A "teaser" image appears between the author and affiliation
%% information and the body of the document, and typically spans the
%% page.
% \begin{teaserfigure}
%   \includegraphics[width=\textwidth]{sampleteaser}
%   \caption{Seattle Mariners at Spring Training, 2010.}
%   \Description{Enjoying the baseball game from the third-base
%   seats. Ichiro Suzuki preparing to bat.}
%   \label{fig:teaser}
% \end{teaserfigure}

% \received{20 February 2007}
% \received[revised]{12 March 2009}
% \received[accepted]{5 June 2009}

%%
%% This command processes the author and affiliation and title
%% information and builds the first part of the formatted document.
\maketitle

\section{Introduction}
\label{sec:intro}

Deploying and improving conversational AI hinges on evaluation under realistic user interactions. Expert human-annotated utterances, while authoritative, is costly and difficult to scale for systematic evaluation. At the same time, the instruction-tuning of models for production systems demands large volumes of authentic user–assistant exchanges. To meet both needs, practitioners increasingly rely on \emph{user proxy agents}, namely LLMs prompted to emulate specified human personas, to synthesize user utterances for automated testing and post-training at scale.

In practice, user proxies are now used far beyond toy role-play: they drive regression testing for assistant updates \cite{wang2025configurablemultiagentframeworkscalable}, evaluate tool-use and grounding behaviors \cite{zhang2025hammerbenchfinegrainedfunctioncalling}, generate synthetic dialogues for post-training \cite{naous2025flippingdialoguetrainingevaluating, hathidara2025disambiguationcentricfinetuningmakesenterprise}, and enable stress tests that are infeasible with human participants \cite{lynch2025agenticmisalignmentllmsinsider, zhang2025searchingprivacyrisksllm}. However, naive ``act-as-a-user'' prompting often yields verbose, overly cooperative utterances that diverge from real user behavior \cite{zhou-etal-2024-real}, indicating that unrealistic proxies can bias downstream evaluations and introduce distribution shift when used for data generation. This motivates a central question: \emph{How can we rigorously measure the human-likeness of a user proxy, independently of downstream task success?} Here, we define a \emph{user} as a human interlocutor participating in a specific conversational or task-oriented dialogue.

% In practice, user proxies are now used far beyond toy role-play: they drive regression testing for assistant updates, evaluate tool-use and grounding behaviors, generate synthetic dialogues for post-training, and support large-scale stress tests that are infeasible with human participants. This shift makes user-proxy quality a \emph{benchmarking} problem: if the proxy systematically differs from real users (e.g., overly verbose, overly helpful, or unrealistically persistent), downstream evaluations can become biased and training data can drift away from genuine user distributions. As a result, improving conversational AI increasingly requires not only better assistants, but also reliable ways to \emph{measure} whether a user proxy is realistic enough for the intended evaluation or data-generation setting.

% Yet the naive prompting of a single LLM to act as user simulator often yields verbose, overly cooperative behavior that diverges from real user speech patterns \cite{zhou-etal-2024-real}. Scalable, practical evaluation of conversational AI therefore depends on \emph{realistic} user simulators, which must themselves be rigorously evaluated for how closely they resemble human users across diverse scenarios. This leads to the central question: \emph{How can we rigorously measure the human-likeness or user-likeness of a user proxy, independently of downstream task success?} Here, we define a \emph{user} as a human interlocutor participating in a specific conversational or task-oriented dialogue.

We introduce \textsc{MirrorBench}, a reproducible and extensible framework for \emph{variance-aware} user-proxy benchmarking  on human-likeness metrics using multiple human reference datasets across diverse conversational settings. \textsc{MirrorBench} evaluates proxies solely on human-likeness and deliberately separates this from task completion. We define the \emph{human-likeness} of a user proxy as the extent to which its generated user utterances match real human user utterances along two complementary axes: (i) \emph{lexical similarity}, operationalized via human-anchored lexical-diversity statistics (e.g., \textsc{MATTR}, \textsc{HD-D}, Yule's $K$), and (ii) \emph{behavioral realism}, operationalized via LLM-judge evaluations that assess whether the proxy's user turns \emph{sound} like a real user in context (e.g., naturalness, tone, appropriateness).

Human-likeness is inherently scenario-dependent: what sounds realistic in open-domain chat may be unrealistic in information-seeking or clarification-heavy interactions. Accordingly, \textsc{MirrorBench} does not aim to capture an unconstrained notion of ``general'' human behavior. Instead, we evaluate \emph{utterance-level realism conditioned on the same task context}, asking whether a proxy produces user turns that resemble those of real users \emph{in that regime}. This framing explicitly decouples user-proxy quality from assistant task success, and treats human-likeness as a measurable property of the \emph{user side} of the interaction under controlled conditions.

To evaluate the realism of user–proxy agents against real-world dialogues, \textsc{MirrorBench} packages four open-source corpora preprocessed for user-proxy evaluation: QULAC \cite{10.1145/3331184.3331265}, ClariQ \cite{aliannejadi2021building}, OASST1 \cite{kpf2023openassistant}, and ChatbotArena \cite{zheng2023judging}. To quantify alignment in lexical diversity with human users, we implement Moving-Average Type-Token Ratio (\textsc{MATTR}) \cite{Covington2010CuttingTG}, \textsc{Yule’s \(K\)} \cite{10.1162/COLI_a_00228}, and a Hypergeometric Distribution Diversity (\textsc{HD-D}) \cite{Carmona636753}.  To assess behavioral realism, we employ a family of LLM-judge metrics, including GTEval \cite{zhu2025dialogueforgellmsimulationhumanchatbot}, Pairwise Indistinguishability (\textsc{PI}) \cite{zheng2023judging}, and Rubric-and-Reason (\textsc{RNR}) \cite{liu2023g}. Taken together, these metrics quantify the realism of the \emph{user} side of the dialogue, explicitly decoupled from assistant capabilities and task success.

Our benchmark reveals that even strong LLMs exhibit systematic gaps from real users, and that these gaps vary substantially across conversational regimes. Across datasets, we observe a recurring realism-diversity tension: proxies that score highly on judge-based behavioral realism can still under-match human lexical/distributional characteristics in clarification-centric settings, motivating the use of complementary metric families. We also find that absolute judge scores and, in some cases, fine-grained orderings can shift with the judge choice, reinforcing the importance of transparent multi-judge reporting and calibration controls when comparing closely matched proxies.

In summary, our contributions are as follows: (1) a benchmark protocol for evaluating user-proxy agents for human-likeness with human-anchored normalization and calibrated judge metrics; (2) six metrics spanning lexical-diversity (\textsc{MATTR}, \textsc{HD-D}, \textsc{Yule's~$K$}) and judge-based realism (\textsc{GTEval}, \textsc{Pairwise Indistinguishability}, \textsc{Rubric-and-Reason}); (3) a unified preprocessing of four open conversational datasets for user-proxy benchmarking; and (4) an empirical study across six user-proxy LLMs, yielding actionable insights on realism-diversity tensions and judge sensitivity.

\section{Related Work}
\label{sec:related_work}

\paragraph{\textbf{User simulation and user proxies.}}
Simulating users to train and evaluate dialogue agents has a long history in NLP \cite{10.1145/3624918.3629549}. Early user simulators were typically goal-driven or rule-based, using predefined agendas over dialogue acts to approximate human behavior \cite{10.1109/TASL.2008.2012071}. More recently, \emph{LLM-based} user proxies have enabled open-ended interactions that better match real conversational variability, but they also introduce failure modes such as verbosity, role drift, and inconsistent intent adherence. For instance, \cite{wang-etal-2025-know} proposes a User Simulator with Implicit Profiles (USP) that infers latent user traits to improve realism beyond naive role-play prompting. LLM user proxies are increasingly used as evaluation drivers in interactive testbeds (e.g., tool-use and multi-turn assistant evaluation) \cite{lu-etal-2025-toolsandbox, ahmad2025simulatinguserdiversitytaskoriented, hathidara2025disambiguationcentricfinetuningmakesenterprise}, and related systems generate synthetic human--assistant conversations from seed dialogues \cite{zhu2025dialogueforgellmsimulationhumanchatbot}. In parallel, broader agent evaluation frameworks (e.g., AgentBench \cite{agentbench2023} and AgentScope \cite{agentscope2024}) evaluate \emph{agents in environments} but do not isolate the human-likeness of the user role.

\paragraph{\textbf{User-proxy evaluation.}} SimulatorArena \cite{dou-etal-2025-simulatorarena} investigates if profile-conditioned LLM simulators can \emph{replace human judges} for ranking AI assistants, demonstrating strong rank correlation on two narrow task types (math tutoring, document creation). SimUSER \cite{bougie-watanabe-2025-simuser} builds persona-based user agents for recommender system evaluation, measuring non-conversational behavioral alignment (clicks, engagement, etc.) at micro and macro levels. Both works treat the simulator as a downstream evaluation tool rather than as a subject of study in its own right. Critically, neither measures \emph{lexical diversity} of generated user turns nor examines the tension between realism and diversity, a key finding in our work, where judge-based indistinguishability and human-anchored vocabulary alignment can diverge substantially across datasets and models.

\paragraph{\textbf{LLMs as judges.}}
LLM-as-a-judge evaluation prompts a strong model to rate or compare dialogue quality, reducing reliance on reference answers or large-scale human annotation \cite{liu-etal-2023-g,li-etal-2023-halueval}. Systems such as MT-Bench/ChatbotArena \cite{zheng2023judging}, G-Eval \cite{liu-etal-2023-g}, and rubric-driven evaluators such as Prometheus \cite{kim2024prometheus} have enabled scalable, flexible assessments by operationalizing custom criteria in natural language prompts. However, most LLM-judge setups focus on evaluating \emph{assistant} outputs, whereas our setting repurposes judge-based evaluation to assess the \emph{user} side and pairs it with explicit calibration controls (HH/PP) and complementary lexical/distributional metrics to better contextualize score magnitudes across datasets and model families.

\section{MirrorBench}
\label{sec:mirrorbench}

\begin{figure*}[t]
    \centering
    \includegraphics[width=0.8\textwidth, trim = 0cm 0cm 0cm 0cm, clip]{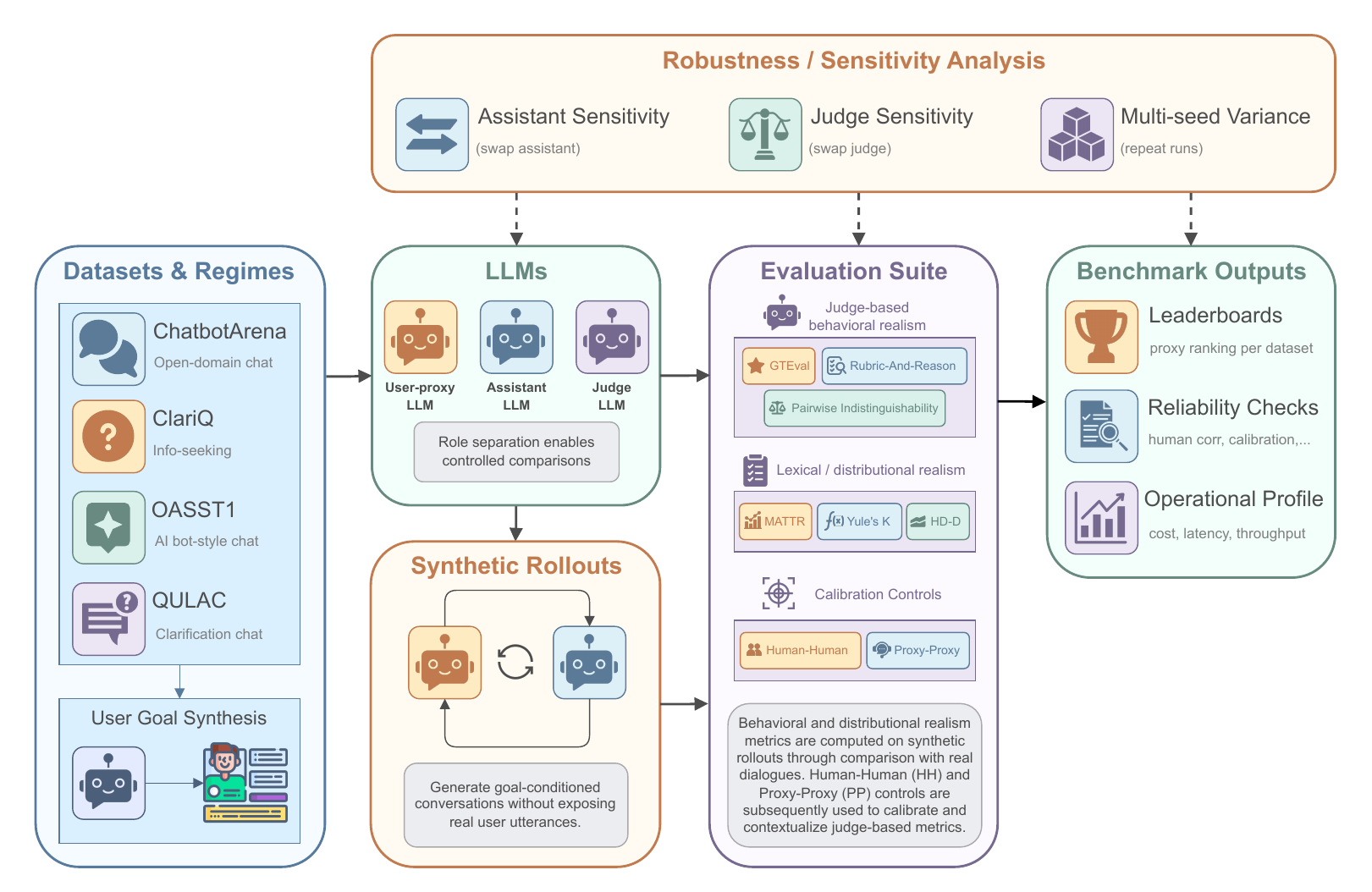}
    \caption{MirrorBench benchmark pipeline, spanning dataset preparation, synthetic rollouts, the evaluation suite, and an overlay of robustness/sensitivity analysis.}
    \label{fig:benchmark_pipeline}
\end{figure*}

This section describes the MirrorBench evaluation methodology for evaluating user-proxy LLMs across diverse conversational datasets. Figure \ref{fig:benchmark_pipeline} summarizes the MirrorBench pipeline. We start from diverse conversational datasets and synthesize user goals to drive controlled, goal-conditioned rollouts between a user-proxy LLM and an assistant LLM under explicit role separation. The resulting transcripts are scored with a combined evaluation suite of judge-based behavioral realism metrics and lexical/distributional metrics, with Human–Human (HH) and Proxy–Proxy (PP) controls used to contextualize judge scores. Finally, we run robustness analyses, swapping assistants or judges and repeating runs across seeds, to quantify sensitivity and variance before reporting benchmark outputs such as leaderboards, reliability checks, and operational profiles. Detailed system architecture is described in Appendix \ref{appsec:system}.

\subsection{Benchmark Protocol}
\label{subsec:mirrorbench->benchmark_proto}

We formalize evaluation of a \emph{user–proxy agent} \(\theta_u\) on a domain-agnostic dialogue dataset \(D^{\mathrm{ref}}\) with respect to a set of human-likeness metrics \(M\). Let
\[
    %D = \{ d^r_1, d^r_2, d^r_3,..., d^r_n \}
    D^{\mathrm{ref}} \;=\; \{\, d^{\mathrm{ref}}_j \in \mathcal{D}\,\}_{j=1}^{n}
\]
denote a corpus of reference, human-grounded dialogues (where $\mathcal{D}$ is the set of finite dialogues). Each dialogue is a sequence of user/assistant turns,
\[
    d^{\mathrm{ref}}_j \;=\; \bigl[(u^{\mathrm{ref}}_{j,1}, a^{\mathrm{ref}}_{j,1}),\ldots,(u^{\mathrm{ref}}_{j,L_j}, a^{\mathrm{ref}}_{j,L_j})\bigr],
\]
with variable length $L_j \!\ge\! 1$. Here $u^{\mathrm{ref}}_{j,t}$ is a \emph{human} user utterance at turn $t$, and $a^{\mathrm{ref}}_{j,t}$ is the corresponding assistant reply (curated by a human expert or generated by an LLM in the source dataset).

\paragraph{\textbf{Goal specification.}}
In some datasets, each reference dialogue $d^{\mathrm{ref}}_j$ includes an explicit user-goal annotation $g_j \in \mathcal{G}$. When no goal is provided, we derive one using an auxiliary goal generator $\theta_{\mathrm{goal}} : \mathcal{D} \to \mathcal{G}$ applied to $d^{\mathrm{ref}}_j$:
\[
g_j \;=\;
\begin{cases}
\text{given goal attached to } d^{\mathrm{ref}}_j, & \text{if available},\\[2pt]
\theta_{\mathrm{goal}}\!\bigl(d^{\mathrm{ref}}_j\bigr), & \text{otherwise}.
\end{cases}
\]
The goal specification is used to condition $\theta_u$ when synthesizing user utterances; if $D^{\mathrm{ref}}$ already provides such annotations, the goal-inference step is omitted.

\paragraph{\textbf{Synthetic rollouts.}}
To compare a user–proxy agent’s behavior against human references, we synthesize proxy–assistant dialogues by rolling out the user–proxy \(\theta_u\) against an assistant model \(\theta_a\). For each reference dialogue \(d^{\mathrm{ref}}_j\) with user goal \(g_j\), we condition \(\theta_u\) on \(g_j\) and the accumulated dialogue history to produce user turns \(\hat{u}_{j,t}\), while \(\theta_a\) produces contextual assistant replies \(\hat{a}_{j,t}\). Here, we condition \(\theta_a\) on \(d^{\mathrm{ref}}_j\) such that the synthetic rollout trajectory follows the same path as real one. This is a deliberate design choice: anchoring the assistant to the reference history keeps the synthetic rollout grounded on the same conversational trajectory as the reference dialogue, ensuring that the proxy and the real user are compared under identical context. This yields a synthetic transcript
\[
\hat{d}_j \;=\; \bigl[(\hat{u}_{j,1}, \hat{a}_{j,1}),\ldots,(\hat{u}_{j,L_j}, \hat{a}_{j,L_j})\bigr],
\]
which we compare with the corresponding reference dialogue \(d^{\mathrm{ref}}_j\) to score human-likeness of the user-proxy. Importantly, our evaluation targets the user side \(\{\hat{u}_{j,t}\}_{t=1}^{L_j}\) only: we consider the behavior, tone, and style of the proxy's utterances, and do not score assistant response content.

\paragraph{\textbf{Task formalization.}}
Let \(T\) denote the fully specified, realism-evaluation task for the \emph{human-likeness} of a user-proxy agent \(\theta_u\) over the reference corpus \(D^{\mathrm{ref}}\). The task encapsulates all configuration required for execution and metric aggregation, ensuring that the evaluation process is reproducible and self-contained.

\textbf{Inputs.} The inputs to \(T\) include a reference dataset \(D^{\mathrm{ref}}\), an optional goal generator \(\theta_{\mathrm{goal}}\) (used when no explicit goal is available), the user-proxy agent under test \(\theta_u\), an assistant model \(\theta_a\) used for synthetic rollouts, and a set of human-likeness metrics \(M\).

\textbf{Outputs.} The evaluation can be expressed as a composite executable function
\[
    R_{\theta_u}
    \;=\;
    \Psi_T\!\bigl(\theta_u,\, D^{\mathrm{ref}},\, M;\, \theta_a,\, \theta_{\mathrm{goal}}\bigr),
\]
where \(\Psi_T\) denotes the instantiated evaluation pipeline. This pipeline performs rollout generation by simulating synthetic dialogues \(\hat{D} = \{\hat{d}_j\}_{j=1}^n\) between \(\theta_u\) and \(\theta_a\) under the same dialogue structures and goals as \(D^{\mathrm{ref}}\). It then applies the metric suite \(M\) to each dialogue pair \((\hat{d}_j, d^{\mathrm{ref}}_j)\), and aggregates the resulting scores across dialogues.  Given a user-proxy agent \(\theta_u\), the output \(R_{\theta_u}\) is a structured evaluation record containing per-sample metric values, aggregated results with confidence intervals, and auxiliary metadata such as random seeds,  model identifiers, and telemetry reports. We also report metric-level calibration and reliability diagnostics that quantify metric stability and judge variability.

\subsection{Datasets \& Tasks}
\label{subsec:mirrorbench->datasets}

%% TODO: Provide appendix ref for dataset cleaning

We evaluate user proxies across four open-source conversational datasets spanning diverse domains and interaction patterns, collectively providing \textbf{795} conversations with real human user turns, enabling direct comparison between proxy-generated and real user behavior. 

% More details about the preprocessing of datasets are appended to Appendix \ref{app:benchmark_datasets} and Table \ref{tab:datasets}.

 %%%%%%%%%%%%%%%% DATASET TABLE %%%%%%%%%%%%%%%%

\begin{table}[h]
\caption{Datasets used for the evaluation. All datasets are open-source and include real human user utterances.}
\label{tab:datasets}
% \vskip 0.15in
% \begin{center}
% \begin{small}
\begin{tabular}{lcc}
\toprule
Dataset & Samples & Avg Turns \\
\midrule
ChatbotArena \cite{zheng2023judging} & 195 & 2.5 \\
\midrule
ClariQ \cite{aliannejadi2021building} & 200 & 7.0 \\
\midrule
OASST1 \cite{kpf2023openassistant} & 200 & 3.3 \\
\midrule
QULAC \cite{10.1145/3331184.3331265} & 200 & 2.0 \\
\bottomrule
\end{tabular}
% \end{small}
% \end{center}
\vskip -0.1in
\end{table}

%%%%%%%%%%%%%%%% DATASET TABLE %%%%%%%%%%%%%%%%

In Table \ref{tab:datasets}, notice that the number of samples we use for benchmarking does not exhaust all the samples of the underlying open-source datasets. Instead, we perform \emph{stratified} sampling over the original datasets to extract the final evaluation datasets.

For each dataset, we first normalize conversations into alternating user-assistant turn sequences, retaining only English dialogues with at least two turns. We then define dataset-specific stratification keys to ensure broad coverage: OASST1 \& ChatbotArena conversations are bucketed by the number of user turns (short, medium, long), QULAC entries are grouped by topic type \& facet category, and ClariQ dialogues are distributed across topic buckets \& clarification-pair counts. Within each stratum, we allocate samples proportionally to the population size while enforcing a minimum-per-stratum threshold to prevent underrepresentation. Using a fixed random seed for reproducibility, we sample up to 200 samples per dataset, yielding a balanced benchmark that spans diverse conversational structures, topic domains, and interaction patterns without oversampling any single mode.
 
For each conversation in all four datasets, we generate a user goal description using an auxiliary LLM ($\theta_g$ defined in \S~\ref{subsec:mirrorbench->benchmark_proto}) that summarizes the user's intent, behavior, tone, and persona based on the real conversation. This description serves as the initialization prompt for user proxies during evaluation.

\subsection{Metrics}
\label{subsec:mirrorbench->metrics}

\textsc{MirrorBench} quantifies the human-likeness of user proxies with two metric families: (i) \emph{lexical-diversity metrics} measure vocabulary richness and repetition patterns (\S~\ref{subsubsec:mirrorbench->metrics->lexical}), and (ii) \emph{judge-based realism metrics} measure behavioral realism (Section~\ref{subsubsec:mirrorbench->metrics->realism}). All metrics are \emph{human-anchored}: proxy scores are compared against the empirical distribution of real-user utterances from the \emph{same} dataset and under same tokenization, yielding interpretable, calibrated results (implementation details in Appendix \ref{appsubsec:system->metric_impl}).

\subsubsection{Lexical Diversity Metrics}
\label{subsubsec:mirrorbench->metrics->lexical}

Lexical-diversity metrics capture how varied a user proxy's word choices are and how often they repeat tokens. Because raw values are sensitive to sequence length, domain, and tokenization, we normalize (\(z\)-score) proxy scores with respect to the human distribution on the same dataset.

\paragraph{\textbf{Moving Average Type-Token Ratio (MATTR)}}
The classical type–token ratio (TTR) \cite{chotlos1944iv,10.5555/539986} decreases as sequence length grows. The moving-average TTR (MATTR) mitigates this length bias by averaging TTR over a sliding window of width \(w\) \cite{Covington2010CuttingTG}. Let \(\text{types}(\cdot)\) return the set of distinct token types in its argument. For a token sequence \(\mathbf{t}=(t_1,\ldots,t_N)\) and window size \(w\le N\),
\begin{equation}
\mathrm{MATTR}_w(\mathbf{t})
\;=\;
\sum_{i=1}^{N-w+1}
\frac{\bigl\lvert \mathrm{types}(t_i,\ldots,t_{i+w-1}) \bigr\rvert}{w (N-w+1)},
\label{eq:mattr}
\end{equation}
where \(\mathrm{types}(\cdot)\) returns the set of distinct token types in the window. We use \(w=50\) by default unless stated otherwise. Note that for \emph{Qulac} dataset, human reference sequences are very short, so MATTR degrades to plain TTR; lexical diversity scores for this dataset should be interpreted accordingly.

\paragraph{\textbf{Yule's K}}
Yule's characteristic constant \(K\) \cite{10.1162/COLI_a_00228} summarizes repetitiveness from the token-frequency spectrum. Let \(V_i\) be the number of types occurring exactly \(i\) times, and \(N=\sum_{i\ge 1} i V_i\) the token count. Then
\begin{equation}
K(\mathbf{t}) \;=\; 10^{4}\,\frac{\sum_{i\ge 1} i^{2} V_i \;-\; N}{N^{2}}.
\label{eq:yules_k}
\end{equation}
Lower \(K\) indicates richer, less repetitive text; higher \(K\) indicates greater repetition. The factor \(10^{4}\) provides a convenient numeric range.

\paragraph{\textbf{Hypergeometric Distribution Diversity (HD-D)}}
HD-D \cite{Carmona636753} estimates vocabulary diversity via hypergeometric sampling, offering robustness to length variation. It approximates the expected number of \emph{distinct} types observed in a sample of size \(s\) drawn \emph{without replacement} from a token sequence \(\mathbf{t}=(t_1,\ldots,t_N)\). For a type \(i\) with frequency \(f_i\) in \(\mathbf{t}\), the probability that the sample contains at least one instance of that type is
\[
P(\text{type}\ i\ \text{in sample}) \;=\; 1 \;-\; \frac{\binom{N-f_i}{\,s\,}}{\binom{N}{\,s\,}},
\quad 1\le s \le N,
\label{eq:hdd_prob}
\]
and the HD-D score is their normalized sum:
\begin{equation}
\mathrm{HD\text{-}D}_s(\mathbf{t}) \;=\; \frac{1}{s}\sum_{i=1}^{V}
\left( 1 \;-\; \frac{\binom{N-f_i}{\,s\,}}{\binom{N}{\,s\,}} \right),
\label{eq:hdd}
\end{equation}
where \(V=\bigl|\text{types}(\mathbf{t})\bigr|\). Following the VOCD-D convention, we use \(s=42\) by default \cite{McCarthy2010VOCDD}.

\paragraph{\textbf{Human-Anchored Z-Score Normalization}}
Raw lexical-diversity scores are dataset-dependent; we therefore report human-anchored $z$-scores computed on the \emph{same} split and tokenizer. Let $\mathcal{H}=\{\mathbf{t}_1^{\mathrm{ref}},\ldots,\mathbf{t}_n^{\mathrm{ref}}\}$ be human user-side token sequences, where each $\mathbf{t}_i^{\mathrm{ref}}$ concatenates the human user turns of one reference dialogue. For any $m\in\{\mathrm{MATTR}_w, K, \mathrm{HD\text{-}D}_s\}$, let
\(
\mu_m^{\text{hum}} = \frac{1}{|\mathcal{H}|} \sum_{\mathbf{t} \in \mathcal{H}} m(\mathbf{t})
\)
and
\(
\sigma_m^{\text{hum}} = \sqrt{\frac{1}{|\mathcal{H}| - 1} \sum_{\mathbf{t} \in \mathcal{H}} \bigl(m(\mathbf{t}) - \mu_m^{\text{hum}}\bigr)^2 }.
\)
During evaluation, each rollout episode produces a synthetic proxy-assistant dialogue, which we denote by \(\hat{d}_i\) (\S~\ref{subsec:mirrorbench->benchmark_proto}). From \(\hat{d}_i\), we extract only the \emph{proxy user} turns (i.e., the outputs of the user–proxy agent \(\theta_u\)) and concatenate them in temporal order to obtain a proxy user token sequence \(\hat{\mathbf{t}}_i\). This \(\hat{\mathbf{t}}_i\) is the proxy-side analogue of the human sequence \(\mathbf{t}_i^{\text{ref}}\) in \(\mathcal{H}\).  We then assign that rollout episode a human-anchored standardized lexical score:
\[
z_m(\hat{\mathbf{t}}_i)
\;=\;
\frac{m(\hat{\mathbf{t}}_i) - \mu_m^{\text{hum}}}{\sigma_m^{\text{hum}}}.
\label{eq:zscore}
\]
% \begin{equation}
% z_{m_k}(\hat{d_i}) = \frac{m_k(\hat{d_i}) - \mu_{m_k}^{\text{hum}}}{\sigma_{m_k}^{\text{hum}}}
% \label{eq:zscore}
% \end{equation}
%When \(\sigma_m^{\text{hum}} = 0\) (i.e., all human episodes achieve the same value under metric \(m\)), we set all \(z_m = 0\) by convention.
%\(z_m = 0\) indicates proxy behavior indistinguishable from real users
By design, \(z_m \approx 0\) indicates that the proxy's lexical behavior matches the human mean for metric \(m\). The sign of \(z_m\) is metric-dependent: for \(\mathrm{MATTR}_w\) and \(\mathrm{HD\text{-}D}_s\), \(z_m > 0\) corresponds to \emph{greater} lexical diversity than the human average; for Yule's \(K\), \(z_m > 0\) corresponds to \emph{more} repetition (i.e., \emph{lower} diversity).  For reporting at the dataset level, we aggregate across all rollout dialogues \(\{\, \hat{d}_i \in \hat{D}\,\}_{i=1}^{n}\)
executed under a fixed configuration. Let \(\{z_m(\hat{\mathbf{t}}_1), \ldots, z_m(\hat{\mathbf{t}}_n)\}\) be the per-episode standardized scores for metric \(m\). We report their sample mean \(\bar{z}_m\) together with a two-sided \(95\%\) confidence interval derived from the Student-\(t\) distribution:
\begin{equation}
\mathrm{CI}_{0.95}(m)
\;=\;
\bar{z}_m
\;\pm\;
t_{0.975,\,n-1}\,
\frac{s_m}{\sqrt{n}},
\label{eq:ci}
\end{equation}
where \(s_m\) is the unbiased sample standard deviation, and \(t_{0.975,\,n-1}\) is the \(97.5\)th percentile of the \(t\)-distribution with \(n-1\) degrees of freedom. These aggregated statistics, namely \(\bar{z}_m\), \(s_m\), and \(\mathrm{CI}_{0.95}(m)\), are the quantities that \textsc{MirrorBench} reports for downstream comparison across user proxies, datasets, and seeds.

\subsubsection{Judge-Based Realism Metrics}
\label{subsubsec:mirrorbench->metrics->realism}

Lexical diversity alone cannot capture whether a simulated user \emph{feels human-like}. Human-likeness also depends on discourse phenomena such as tone, politeness, hesitations, or style.  \textsc{MirrorBench} therefore includes \emph{judge-based realism metrics}: LLM evaluators that score proxy behavior against human references along higher-level dimensions. Each judge call uses a chain-of-thought style prompting strategy \cite{10.5555/3600270.3602070}: the judge is instructed to reason privately and then produce a final structured verdict. To improve robustness, each judge is executed with a self-consistency parameter \(c \ge 1\), which repeats the judgment \(c\) times under the same prompt and aggregates the resulting scores.

\paragraph{\textbf{GTEval (Relative Realism Scoring)}}
GTEval \cite{zhu2025dialogueforgellmsimulationhumanchatbot} compares a proxy-generated user transcript against a human reference and returns a similarity score in \([0,1]\), where higher values indicate that the proxy is judged to behave more like the human reference. The judge model (with \(c=1\)) is shown the pair \((\hat{d}_j, d_j^{\mathrm{ref}})\) and asked to assess stylistic and behavioral similarity (e.g., tone, naturalness). Let \(\mathrm{J}^{(r)}(\cdot) \in [0,1]\) denote the scalar verdict returned by the judge on repetition \(r\). The GTEval score for episode \(j\) is
\begin{equation}
s^{(\mathrm{GTEval})}_j
\;=\;
\frac{1}{c} \sum_{r=1}^{c}
\mathrm{J}^{(r)}\!(
\hat{d}_j,\,
d_j^{\mathrm{ref}}
),
\label{eq:gteval}
\end{equation}
where each \(\mathrm{J}^{(r)}\) corresponds to an independent call to the judge with a different random seed.

\paragraph{\textbf{Pairwise Indistinguishability (PI)}}
Here, judge LLM is given two anonymized user-assistant conversations, and asked to choose one in which the user sounds more like a human. For each sample, we present one proxy conversation \(\hat{d}_j\) and one human reference \(d_j^{\mathrm{ref}}\) in random order, unlabeled except as \(\mathrm{A}\) or \(\mathrm{B}\). The judge returns a verdict \(v_{j,r} \in \{\mathrm{A}, \mathrm{B}, \mathrm{Tie}\}\) for judgment round \(r\). Let \(p_{j,r} \in \{\mathrm{A}, \mathrm{B}\}\) be the position where the proxy was shown for that comparison (randomized per judgment). With \(c\) repeated judgments (default \(c=3\)), we define the per-episode win rate
\begin{equation}
w^{(\mathrm{PI})}_j \;=\;
\frac{1}{c} \sum_{r=1}^{c} (
\mathds{1}[v_{j,r} = p_{j,r}] \;+\;
0.5\,\mathds{1}[v_{j,r} = \mathrm{Tie}]).
\label{eq:pi_win}
\end{equation}
Aggregated win rate is summarized by the mean \(\bar{w}^{(\mathrm{PI})} = \frac{1}{n} \sum_{j=1}^{n} w^{(\mathrm{PI})}_j\) and its centered form \(\Delta w^{(\mathrm{PI})} = \bar{w}^{(\mathrm{PI})} - 0.5\). Here, \(\Delta w > 0\) indicates the judge tends to prefer the proxy user over the human baseline, \(\Delta w < 0\) favors real users, and \(\Delta w \approx 0\) indicates that the proxy is effectively \emph{indistinguishable} from the human baseline.

\paragraph{\textbf{Rubric-and-Reason (RNR)}}
RNR adapts rubric-style LLM evaluation to judge human-likeness without requiring a paired human reference. The judge sees only the proxy-side user transcript \(\hat{d}_j\) and a rubric that defines realism along dimensions such as style, behavior, and tone. For a given rubric, the judge returns a binary verdict \(v^{(r)} \in \{\mathrm{YES}, \mathrm{NO}\}\) on repetition \(r\), which we map to
\[
\hat{s}_{\mathrm{RNR}}^{(r)}(\hat{d}_j) \;=\;
\begin{cases}
1.0, & \text{if } v^{(r)} = \mathrm{YES},\\
0.0, & \text{if } v^{(r)} = \mathrm{NO}.
\end{cases}
\]
We repeat each judgment \(c=2\) times. The aggregate RNR score for sample \(j\) is
\begin{equation}
s_{\mathrm{RNR}}(\hat{d}_j)
\;=\;
\frac{1}{c}
\sum_{r=1}^{c}
\hat{s}_{\mathrm{RNR}}^{(r)}(\hat{d}_j).
\label{eq:rnr}
\end{equation}
Unlike GTEval and PI, which are explicitly comparative, RNR is reference-free and produces an absolute realism score for the proxy alone. Because this absolute score depends on the judge model and the elaborated rubric, it is interpreted \emph{relatively}: higher \(s_{\mathrm{RNR}}(\hat{d}_j)\) means the proxy is behaving more like a real user under the fixed judge and rubric set. Optionally, we also evaluate the human reference \(d_j^{\mathrm{ref}}\) under the same procedure to obtain \(s_{\mathrm{RNR}}(d_j^{\mathrm{ref}})\), which serves as an empirical upper bound (useful for calibration).

\begin{figure*}[t]
    \centering
    \includegraphics[width=0.8\textwidth, trim = 0cm 0cm 0cm 0cm, clip]{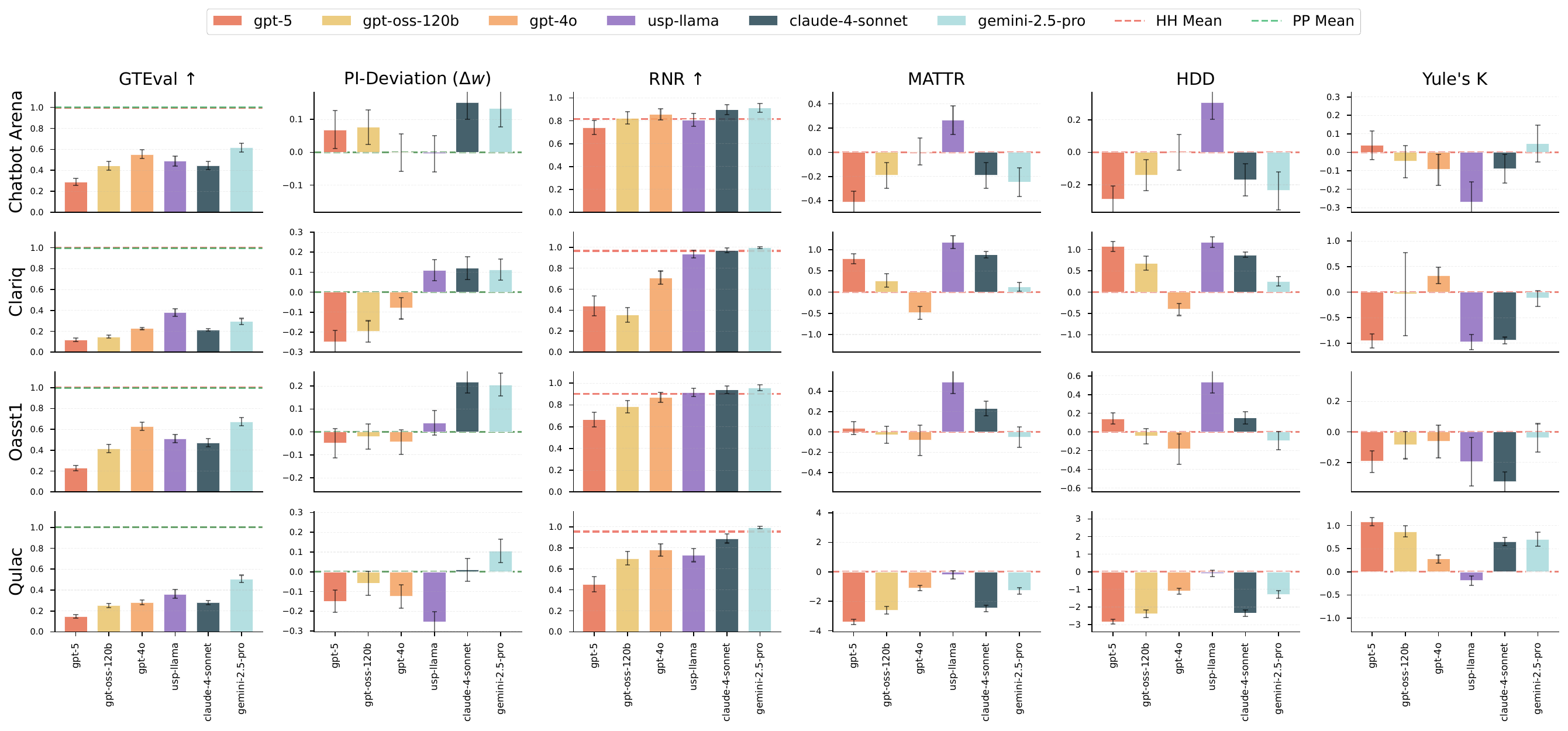}
    \caption{\textbf{Human-likeness of six user-proxy LLMs across four datasets.} Higher is better for judge-based metrics (GTEval, PI, RNR). Lexical-diversity metrics are \(z\)-scored to human baselines (0 is best). We fix the judge to Claude-4-Sonnet and the assistant to GPT-4o.}
    \label{fig:user_proxy_results}
\end{figure*}

\paragraph{\textbf{Calibration via HH and PP Controls}}
Judge models can exhibit systematic bias (e.g., favoring verbosity, excessive politeness, or self-preference). To expose this bias and to place GTEval and PI scores on an interpretable scale, we optionally compute two control conditions:
\begin{itemize}
\item \textbf{Human-Human (HH):} Compare a human conversation to itself,
\(s_{\mathrm{HH},j} = \mathrm{Judge}(d_j^{\mathrm{ref}}, d_j^{\mathrm{ref}})\).
This approximates the judge's effective ceiling for genuine human behavior on episode \(j\).
\item \textbf{Proxy-Proxy (PP):} Compare a proxy conversation to itself,
\(s_{\mathrm{PP},j} = \mathrm{Judge}(\hat{d}_j, \hat{d}_j)\).
This estimates how the judge scores the proxy in isolation.
\end{itemize}
For GTEval, both \(s_{\mathrm{HH},j}\) and \(s_{\mathrm{PP},j}\) are typically expected to be close to 1. For PI, a self comparison should yield an ideal effective win rate near 0.5.  We calculate their empirical means across all \(n\) samples:
\(
\mu_{\mathrm{HH}} = \tfrac{1}{n}\sum_{j=1}^{n} s_{\mathrm{HH},j},
\)
\(
\mu_{\mathrm{PP}} = \tfrac{1}{n}\sum_{j=1}^{n} s_{\mathrm{PP},j}.
\)
For PI Metric, we also calculate calibrated score \(s_{\mathrm{cal}}\). Let \(s_{\mathrm{raw}}\) denote the uncalibrated proxy score of interest (e.g. \(\bar{w}^{(\mathrm{PI})}\)). We define a calibrated \([0,1]\) score via an affine rescaling between the PP and HH anchors:
\begin{equation}
s^{PI}_{\mathrm{cal}} \;=\;
\mathrm{clip}\!\left(
\frac{s_{\mathrm{raw}} - \mu_{\mathrm{PP}}}
{\max\!\bigl(\epsilon,\; \mu_{\mathrm{HH}} - \mu_{\mathrm{PP}}\bigr)},
\;0,\;1
\right),
\label{eq:calibration_final}
\end{equation}
where \(\epsilon = 10^{-6}\) prevents division by zero and \(\mathrm{clip}(\cdot,0,1)\) bounds the result. Under this normalization, \(s_{\mathrm{cal}} \approx 0\) means the proxy is no better than its own PP baseline, while \(s_{\mathrm{cal}} \approx 1\) means the proxy is judged as indistinguishable from human under the same judge.

\section{Experiments \& Results}
\label{sec:results}

We evaluate multiple user-proxy LLMs across four diverse conversational datasets, ChatbotArena, ClariQ, OASST1, and QULAC, to quantify human-likeness using both lexical-diversity and LLM-judge realism metrics. Concretely, we report \textsc{MATTR}, \textsc{Yule’s $K$}, and \textsc{HD-D} (human-anchored \(z\)-scores), together with three judge-metrics \textsc{GTEval}, \textsc{PI}, and \textsc{RNR}. Our experiments (i) measure how closely different proxy architectures reproduce human user behavior, (ii) assess the reliability and calibration of LLM-as-judge metrics, and (iii) characterize computational cost, latency, and throughput under large-scale runs.

We compare six LLMs as user proxies: GPT-4o \cite{hurst2024gpt}, GPT-5 \cite{openai_gpt5_system_card_2025_tr}, GPT-OSS-120B \cite{openai2025gptoss120bgptoss20bmodel}, Claude-4-Sonnet \cite{anthropic_claude4_system_card_2025_tr}, Gemini-2.5-Pro \cite{comanici2025gemini25pushingfrontier}, and a specialized open-source user-simulation proxy (USP) based on LLaMA \cite{wang-etal-2025-know}. Unless otherwise noted, the assistant is fixed to GPT-4o and the judge to Claude-4-Sonnet for all primary results, enabling apples-to-apples comparisons across proxies. All the evaluations are performed with a single seed (unless otherwise specified), and we report aggregated scores with 95\% confidence intervals. Judge scores are calibrated, and we include a human correlation check to validate judge trends. We also analyze fine-grained telemetry to make cost/performance trade-offs explicit.

\subsection{Comparison of User-Proxy LLMs}

Figure~\ref{fig:user_proxy_results} compares six user-proxy LLMs across four datasets with a fixed judge (Claude-4-Sonnet) and assistant (GPT-4o). The left three columns of subplots report judge-based realism (\textsc{GTEval}, \textsc{PI~$\Delta w$}, \textsc{RNR}; higher is better); the right three columns report human-anchored lexical diversity (\textsc{MATTR}, \textsc{HD-D}, \textsc{Yule's~K}; best at the human z-score = 0). Dashed red (HH) and green (PP) lines provide calibration controls; error bars denote 95\% CIs. We provide the assistant the reference chat history to anchor the rollout to the same trajectory as the original conversation; while LLM hallucinations can occasionally induce drift, a manual inspection found such divergence in mere $<1\%$ of samples evaluated as part of \ref{fig:user_proxy_results}.

\paragraph{\textbf{Judge realism is consistent across datasets}}
Across GTEval, PI~$\Delta w$, and RNR, \emph{Gemini-2.5-Pro} and \emph{Claude-4-Sonnet} are the most human-like on every dataset, with \emph{GPT-4o} competitive but generally behind, and \emph{GPT-OSS-120B} and \emph{GPT-5} trailing. On ClariQ and QULAC, Claude-4-Sonnet/Gemini-2.5-Pro approach the HH ceiling under RNR, while PI~$\Delta w$ shows clear positive win margins for these models and negative/near-zero deltas for the rest. The agreement among the three judge metrics indicates a stable ordering not driven by any single rubric.

\paragraph{\textbf{Diversity shows strong dataset effects and a realism–diversity tension}}
Lexical diversity diverges from the realism ranking. On \emph{ClariQ}, most notably \emph{Claude-4-Sonnet} and \emph{GPT-5} exceed the human anchor on \textsc{MATTR}/\textsc{HD-D} and exhibit lower \textsc{Yule’s~K}, indicating a more diverse vocabulary than human questioners in this information-seeking regime. In contrast, \emph{QULAC} exhibits a uniform diversity deficit: all proxies fall below the human baseline on \textsc{MATTR} and \textsc{HD-D} (with positive shifts in \textsc{Yule’s~K}), suggesting more templated clarifications than humans. \emph{ChatbotArena} and \emph{OASST1} sit between these extremes with smaller deviations around the human baseline. \textit{Overall, \emph{Gemini-2.5-Pro} \& \emph{GPT-4o} yield the strongest diversity alignment with humans across datasets, whereas \emph{Claude-4-Sonnet} generally tends to overshoot the human baseline.} \emph{GPT-5} and \emph{GPT-OSS-120B} have generally larger diversity difference compared to human baselines.

\paragraph{\textbf{Judge realism and diversity are partially decoupled}}
High judge realism does not guarantee human-level diversity. \emph{Claude-4-Sonnet} and \emph{Gemini-2.5-Pro} lead on \textsc{GTEval}/\textsc{PI}/\textsc{RNR} but under-shoot diversity on \emph{QULAC}, indicating judges favor intent/style over surface variety. Conversely, GPT-4o shows more stable diversity with moderate judge-realism gains, underscoring that diversity alone is insufficient to achieve judge indistinguishability.

\paragraph{\textbf{Uncertainty and robustness.}}
Confidence intervals are generally narrow. When overlaps occur (e.g., \textsc{GTEval} on ChatbotArena), rank differences are small and concur with \textsc{PI}/\textsc{RNR} trends. For lexical-diversity panels, confidence intervals occasionally widen, notably for \textsc{Yule's~K}, due to very short the user-side text (e.g., due to few turns) that increases estimator variance. Overall, Figure~\ref{fig:user_proxy_results} supports three takeaways: (i) \emph{Gemini-2.5-Pro} and \emph{Claude-4-Sonnet} are reliably the most human-like by judge criteria; (ii) diversity depends strongly on dataset regime and proxies often lag on clarification-centric tasks like QULAC; and (iii) realism and diversity capture complementary facets of user-proxy quality, motivating the use of both families of metrics.

\paragraph{\textbf{Specialized open-source proxy (USP-LLaMA)}} We additionally evaluate USP \cite{wang-etal-2025-know}, a LLaMA-based model fine-tuned specifically for user simulation. Notably, despite being a smaller specialized model, USP achieves competitive realism scores, outperforming several much larger general-purpose LLMs on \textsc{GTEval} \& \textsc{RNR} across all four datasets and on PI across two datasets, demonstrating that task-specific fine-tuning can be a cost-effective path toward human-like user simulation. On lexical diversity, USP produces richer vocabulary than real users on ChatbotArena, OASST1, and ClariQ, which suggests that the model has internalized varied language. The main limitation surfaces on QULAC, where \textsc{PI} deviation is notably high, suggesting that clarification-centric tasks remain challenging for specialized proxies. Overall, USP represents a promising direction: fine-tuned open-source proxies can rival proprietary frontier models on realism while being lightweight, though better calibration of lexical diversity and robustness across task types remain open challenges.

\subsection{Judge Sensitivity Analysis}
\label{subsec:results->judge_sensitivity}

\begin{figure}[t]
    \centering
    \includegraphics[width=0.9\columnwidth, trim = 0cm 0.2cm 0cm 0.2cm, clip]{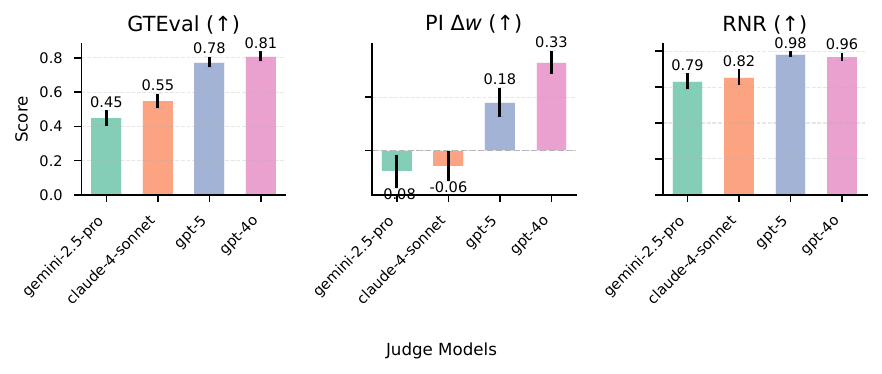}
    \caption{Judge sensitivity of judge-realism metrics on ChatbotArena with assistant \& user-proxy both set to GPT-4o; bars vary the \emph{judge} model. Error bars are 95\% CIs.}
    \label{fig:judge_ablations}
\end{figure}

With the assistant and user-proxy fixed to GPT-4o, Figure~\ref{fig:judge_ablations} shows substantial judge sensitivity on realism scores. On GTEval, scores spread widely ($\approx\!0.45\text{--}0.81$), with GPT-4o as judge yielding the highest value. PI is the most volatile: Gemini-2.5-Pro \& Claude-4-Sonnet produce near-zero or negative win deltas, while GPT-5 \& GPT-4o are clearly positive, suggesting family/self-preference or rubric alignment effects. RNR saturates near the ceiling for GPT-4o \& GPT-5 judges ($\approx\!0.96\text{--}0.98$) and is lower for others ($\approx\!0.79\text{--}0.82$), indicating judge-dependent sensitivity: PI $>$ GTEval $>$ RNR.

These differences imply that conclusions drawn from a single judge can shift both the absolute level and the ordering of models. Since LLM judges can exhibit systematic biases \cite{thakur-etal-2025-judging, sahoo2025quantitativellmjudges}, they may behave either overly generous or overly conservative. Thus in practice, we recommend evaluating with \emph{multiple judges} and applying HH/PP calibration when feasible, or normalizing per-judge before aggregation, to avoid over-interpreting judge-specific biases. Empirically, \emph{Claude-4-Sonnet} behaves as a conservative yet stable judge, yielding non-saturated, well-separated scores across metrics, which is why we use Claude-4-Sonnet as a judge in the results reported in Figure \ref{fig:user_proxy_results}.

\subsection{Human-Judge Correlation}
\label{subsec:results->human_judge_correlation}

Although judge-sensitivity analysis (\S\ref{subsec:results->judge_sensitivity}) helps characterize a judge's behavior on our task, it can still be difficult to interpret, since whether a judge appears ``generous'' or ``conservative'' is itself \emph{task-dependent}. We therefore additionally assess judge reliability via human-judge correlation.

\begin{figure}[t]
    \centering
    \includegraphics[width=0.9\columnwidth, trim = 0cm 1cm 0cm 0cm]{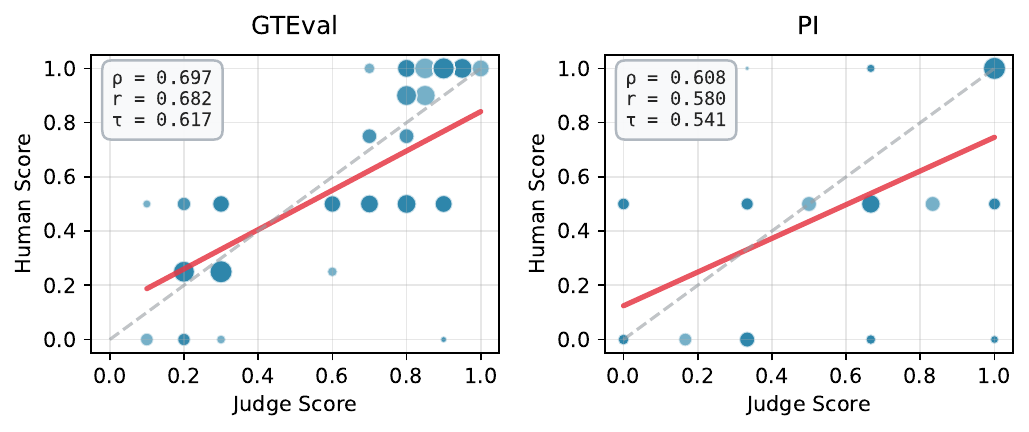}
    \caption{\textbf{Judge–human correlation on ChatbotArena:} Correlation of Claude-4-Sonnet judge scores \& human scores for GTEval and PI, evaluated on \emph{Gemini-2.5-Pro} user-proxy outputs (N=100 per metric). Solid line: linear fit; dashed: identity. Point size encodes local sample density. All correlations p $<$ 0.001.}
    \label{fig:judge_human_correlation}
\end{figure}

We validate judge reliability by correlating Claude-4-Sonnet judge scores with blinded human expert annotations on ChatbotArena. For the analysis shown in Figure~\ref{fig:judge_human_correlation}, we stratified 100 episodes per metric (GTEval, PI) by judge score and conversation length, then compute Spearman's $\rho$ \cite{spearman_coeff}, Pearson's $r$ \cite{10.2307/2346598}, and Kendall's $\tau$ \cite{Kendall1938ANM} using human-judge label pairs. We observe that the GTEval aligns strongly with human judgments, while PI exhibits a moderate correlation. Note that PI is inherently harder to annotate because it requires pairwise comparisons; thus, a moderate correlation still indicates that PI is a reasonably reliable judge metric. Overall, these results suggest that judge scores track human perceptions of user-proxy quality.

To assess whether this finding generalizes, we also conduct a broader correlation study spanning three datasets and three proxy models (50-100 human-annotated episodes per condition), shown in Appendix \ref{app:extended_results}. We observe GTEval $\rho$ ranges from 0.607 to 0.697 and PI $\rho$ from 0.532 to 0.671 (all $p < 0.001$), confirming that judge-human alignment is consistent regardless of dataset domain or proxy model.

\subsection{Assistant Sensitivity and Rank Stability}
\label{sec:assistant_sensitivity}

LLM-as-judge scores can be confounded by the assistant model that the user proxy interacts with. To quantify this \emph{assistant sensitivity}, we rerun evaluation on \emph{ChatbotArena} while holding the judge fixed and varying the assistant among \{GPT-4o, Claude-4-Sonnet, Gemini-2.5-Pro\}. For each user-proxy model, we compute the three judge-based realism metrics (\textsc{GTEval}, \textsc{PI}, \textsc{RNR}) on the resulting rollouts.

\begin{figure}[h]
    \centering
    \includegraphics[width=0.9\columnwidth, trim = 0cm 1cm 0cm 0.5cm]{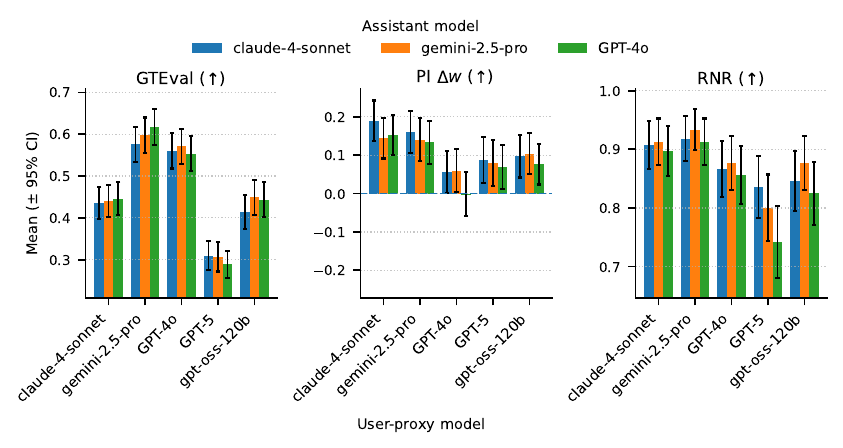}
    \caption{Assistant sensitivity on ChatbotArena (scores). Mean $\pm$ 95\% CI of judge-based realism metrics for each user-proxy when swapping the \emph{assistant model} (color); the judge is fixed to Claude-4-Sonnet.}
    \label{fig:assistant_sensitivity_result}
\end{figure}

\begin{figure}[h]
    \centering
    \includegraphics[width=0.9\columnwidth, trim = 0cm 1cm 0cm 0.5cm]{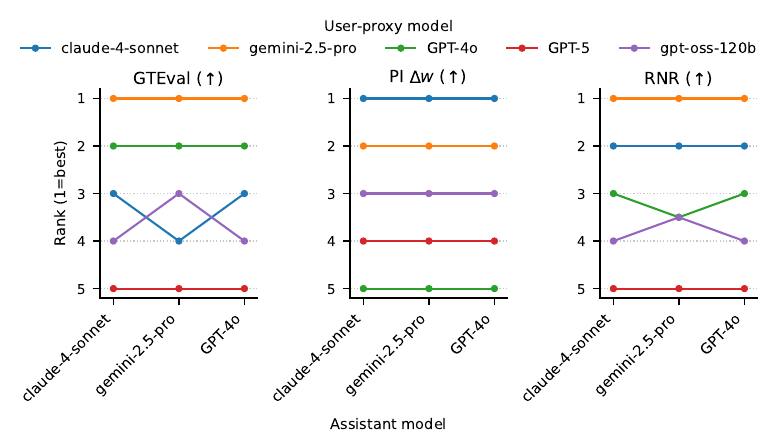}
    \caption{Rank stability under assistant swaps (ChatbotArena).
Per-metric proxy ranks (1 = best) induced by the scores in Fig.~\ref{fig:assistant_sensitivity_result} when varying the assistant model; stable lines indicate robust proxy ordering, while crossings reveal assistant-dependent rankings.}
    \label{fig:assistant_rank_stability}
\end{figure}

Figure~\ref{fig:assistant_sensitivity_result} shows that absolute scores vary only slightly across assistants and are generally insufficient to change the overall ordering of user-proxy models. Figure~\ref{fig:assistant_rank_stability} makes this explicit: ranks are stable (largely parallel trajectories) across assistants, with a single crossover in \textsc{GTEval}. Overall, assistant choice introduces mild variance in judge-based realism scores, but the resulting rankings remain stable most of the time.

\subsection{Multi-Seed Robustness}
\label{subsec:results->multi_seed}

LLM-as-judge metrics can fluctuate across repeated runs due to stochasticity in generation and evaluation. To assess the stability of our judge-based realism metrics, we perform a multi-seed analysis by repeating the benchmark five times with different random seeds for three different user-proxies while keeping the assistant and judge fixed.

\begin{figure}[h]
    \centering
    \includegraphics[width=0.9\columnwidth, trim = 0cm 1cm 0cm 0.5cm]{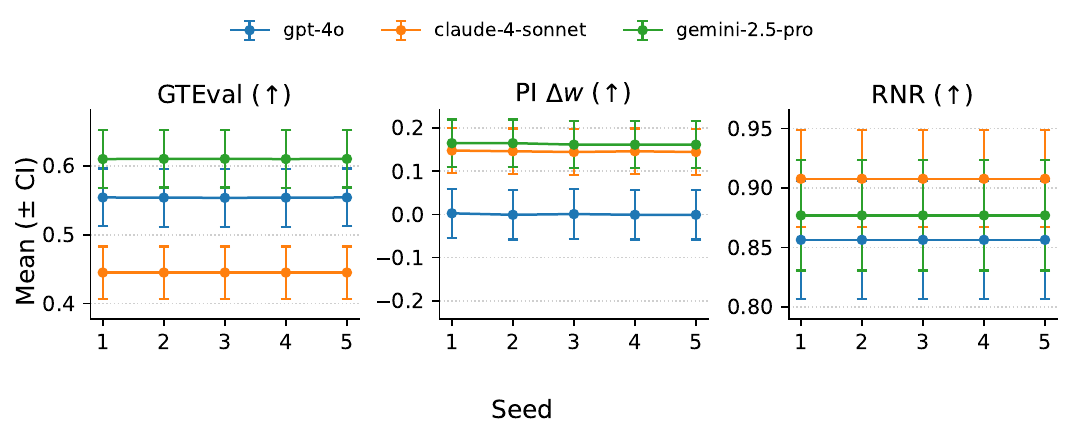}
    \caption{Multi-seed robustness of judge-based realism metrics. For each user-proxy model, points show the per-seed mean score and error bars indicate the corresponding 95\% CI on ChatbotArena; the assistant (gpt-4o) and judge (claude-4-sonnet) are fixed across runs.}
    \label{fig:multi_seed_robustness}
\end{figure}

Figure~\ref{fig:multi_seed_robustness} reports per-seed mean$\pm$95\% CI for \textsc{GTEval}, \textsc{PI}~$\Delta w$, and \textsc{RNR}. Across all three metrics, trajectories are nearly flat, indicating stability of judge-based metrics. The relative ordering of user-proxy models is consistent across seeds, and the small seed-to-seed fluctuations are dominated by within-run uncertainty captured by the error bars. Overall, these results suggest that our conclusions are robust to run-to-run variability in the rollout and judge scoring pipeline.

We also performed extended multi-seed analysis to all four datasets (3 seeds each for ClariQ, OASST1, Qulac), which can be found in Appendix \ref{app:extended_results}. Standard deviation remains $\leq$\,0.016 in all 12 conditions and proxy rankings are stable in 11 of 12 cases, with a single near-tie on OASST1 ($\Delta = 0.004$).

\subsection{Goal-Conditioning Ablation}
\label{subsec:goal_ablation}

The goal description passed to each user proxy is synthesized by an auxiliary LLM from the reference conversation (see \S\ref{subsec:mirrorbench->benchmark_proto}). To quantify how much this conditioning contributes to proxy performance and to assess whether the description leaks task content, we ran a three-condition ablation: \texttt{goal\_full} (full 2--4 sentence intent and persona description, as used for the main experiments), \texttt{goal\_topic} (a short 3-8 word topic phrase, e.g., ``React \texttt{scrollIntoView} debugging''), and \texttt{goal\_none} (no conditioning; equivalent to naive ``act-as-a-user'' prompting). We evaluate GPT-4o and Claude-4-Sonnet proxies on ChatbotArena and ClariQ datasets with a fixed Claude-4-Sonnet judge.

\begin{table}[h]
\centering
\tiny
\caption{Goal-conditioning ablation. GTEval and PI scores for GPT-4o / Claude-4-Sonnet proxies under three conditioning levels; judge fixed to Claude-4-Sonnet.}
\label{tab:goal_ablation}
\begin{subtable}[t]{0.47\columnwidth}
\centering
\tiny
\caption{ChatbotArena}
\setlength{\tabcolsep}{3pt}
\begin{tabular}{lcc}
\toprule
Condition & GTEval & PI \\
\midrule
\texttt{goal\_full}  & 0.554 / 0.446 & 0.499 / 0.652 \\
\texttt{goal\_topic} & 0.548 / 0.378 & 0.539 / 0.641 \\
\texttt{goal\_none}  & 0.156 / 0.018 & 0.296 / 0.022 \\
\bottomrule
\end{tabular}
\end{subtable}
\hfill
\begin{subtable}[t]{0.47\columnwidth}
\centering
\tiny
\caption{ClariQ}
\setlength{\tabcolsep}{3pt}
\begin{tabular}{lcc}
\toprule
Condition & GTEval & PI \\
\midrule
\texttt{goal\_full}  & 0.228 / 0.213 & 0.419 / 0.620 \\
\texttt{goal\_topic} & 0.219 / 0.189 & 0.591 / 0.838 \\
\texttt{goal\_none}  & 0.067 / 0.010 & 0.000 / 0.000 \\
\bottomrule
\end{tabular}
\end{subtable}
\end{table}

Table \ref{tab:goal_ablation} showcases the results, where we observe two insightful findings. First, \texttt{goal\_none} causes near-complete collapse on both datasets: GTEval drops to near-zero and PI collapses to 0 on ClariQ, confirming that intent context is essential. Without it, proxies generate generic, repetitive phrasing with no anchor to a specific conversational objective, which judges immediately identify as non-human. Second, the PI/GTEval divergence under \texttt{goal\_topic} is informative: PI rises \emph{above} \texttt{goal\_full} (e.g., ClariQ Claude: 0.838 vs.\ 0.620), while GTEval \emph{drops} (ClariQ Claude: 0.189 vs.\ 0.213). A short topic phrase produces fluent, human-sounding turns, preserving surface-level indistinguishability, but without the full intent description the proxy drifts from the user's specific objective. This PI-GTEval divergence also serves as evidence against content leakage: if the full goal description were trivially encoding task structure, PI and GTEval would move together rather than in opposite directions. We provide detailed examples to illustrate this in Appendix \ref{app:goal_ablation_examples}.

\subsection{Telemetry, Scaling, and Cost}

\begin{figure}[h]
    \centering
    \includegraphics[width=0.9\columnwidth, trim = 0cm 1.5cm 0cm 1cm]{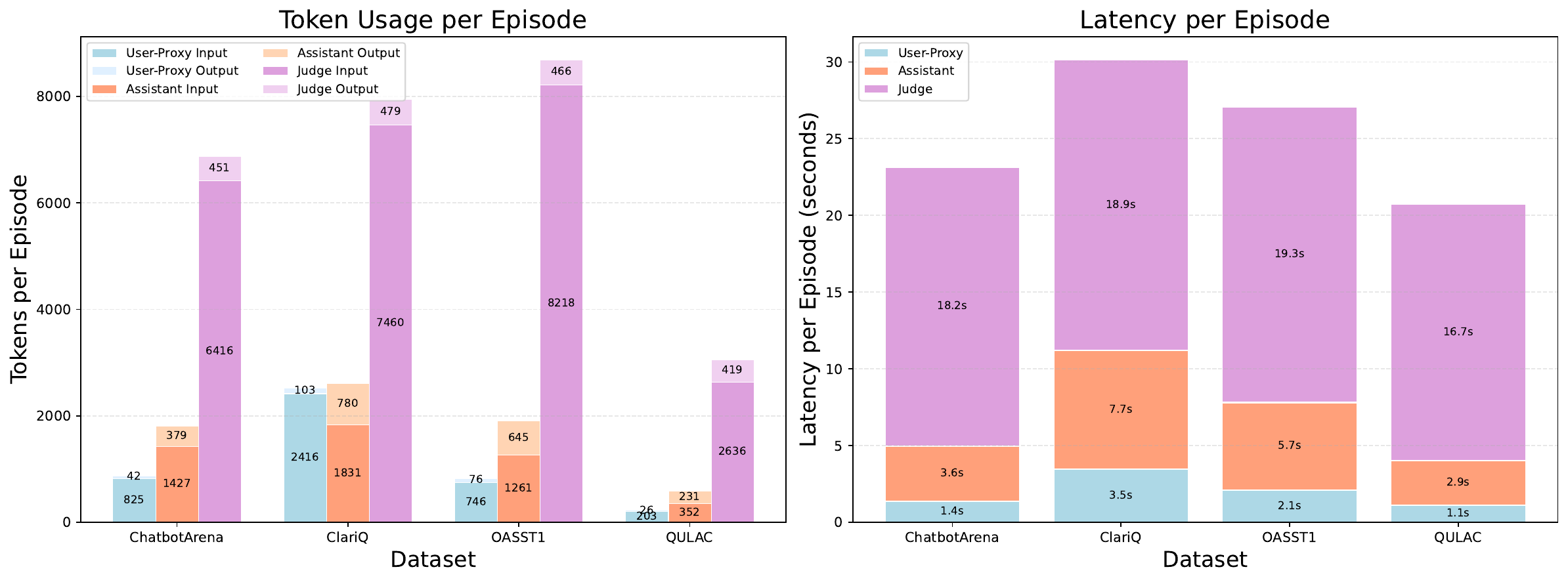}
    \caption{Avg per-sample telemetry for GPT-4o as user-proxy \& assistant, and judged by Claude-4-Sonnet across 4 datasets on 6 metrics (Fig.~\ref{fig:user_proxy_results}). \textbf{Left:} Token usage by role (darker: input, lighter: output). \textbf{Right:} Cumulative latency per episode.}
    \label{fig:one_run_telemetry}
\end{figure}

Based on our experiments, we further provide detailed analysis on the practical side of evaluation. Since most of the models we used are proprietary (incurs cost per million tokens), we provide statistical analysis on cost, latency and trade-off between cost vs performance. Per-sample telemetry (Fig.~\ref{fig:one_run_telemetry}) shows that token usage is dominated by the judge, with the user-proxy \& assistant contributing a smaller but non-negligible share; datasets differ markedly, \emph{OASST1} drives the highest token counts while \emph{ClariQ} yields the largest end-to-end latency.

% \begin{figure}[h]
%     \centering
%     \includegraphics[width=0.7\columnwidth, trim = 0cm 1.5cm 0cm 0cm]{images/scaling_curve_judge_comparison.pdf}
%     \caption{Throughput vs. concurrency for \textsc{GTEval} on ChatbotArena (async backend, cache off); only the \emph{judge} varies. User-proxy and assistant fixed to GPT-4o.}
%     \label{fig:scaling_curve_judge_comparison}
% \end{figure}

\begin{figure}[h]
    \centering
    \includegraphics[width=0.65\columnwidth, trim = 0cm 1.5cm 0cm 0cm]{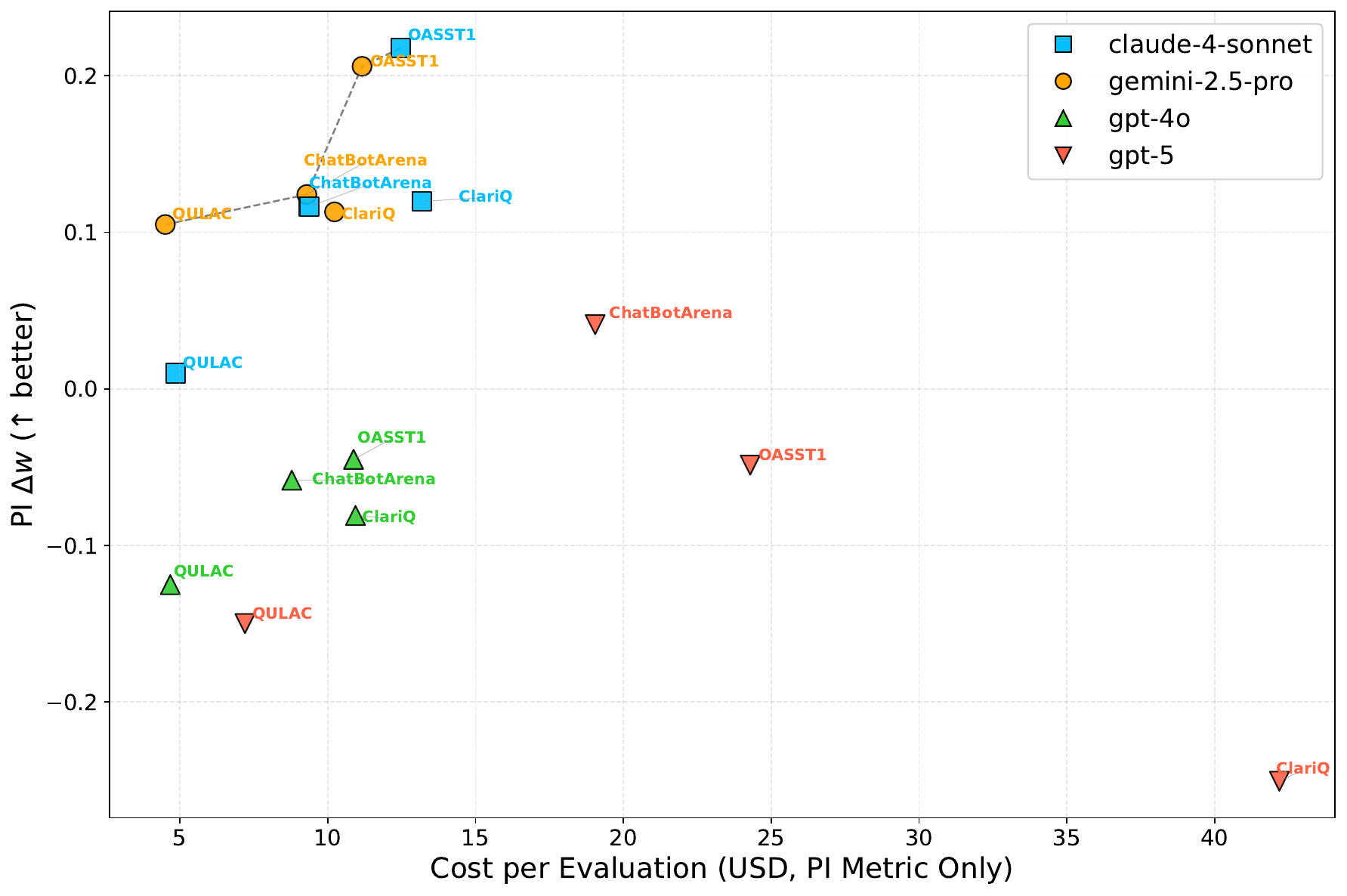}
    \caption{\textbf{Cost-quality trade-off for \textsc{PI}:} cost per evaluation (USD) vs. PI $\Delta w$ ($\uparrow$ better). Judge = Claude-4-Sonnet; assistant = GPT-4o; temperature = 0; cache off. Markers denote user-proxies; labels denote datasets. Dashed line: Pareto frontier. See Table \ref{tab:model-pricing} for model pricing.}
    \label{fig:cost_pi_pareto}
\end{figure}

The cost-quality frontier (Fig.~\ref{fig:cost_pi_pareto}) clarifies trade-offs for PI evaluation: User-proxies Gemini-2.5-Pro and Claude-4-Sonnet offer attractive Pareto points (good PI $\Delta w$ at moderate cost), and GPT-5 generally incurs higher cost with weaker PI gain. Overall, user-proxy Gemini-2.5-Pro, along with Claude-4-Sonnet as judge, provides a balanced choice for large runs.

\section{Conclusion \& Future Work}

\textsc{MirrorBench} provides a reproducible benchmark for evaluating the \emph{human-likeness} of LLM-based user proxies, assessing how closely a proxy's utterances match real human users across conversational tasks, independent of downstream task success. Across four public datasets, we evaluate six user-proxy LLMs using three judge-based realism metrics (\textsc{GTEval}, \textsc{PI}, \textsc{RNR}) together with human-anchored lexical diversity measures (\textsc{MATTR}, \textsc{HD-D}, \textsc{Yule's~K}). The resulting analyses highlight (i) systematic gaps between proxies and human users, (ii) a consistent realism-diversity trade-off that varies by dataset regime, and (iii) measurable sensitivity of absolute judge scores to the judge model, motivating calibration controls (HH/PP) and transparent multi-judge reporting. We also report token, latency, and cost profiles to make the evaluation trade-offs explicit when benchmarking at scale.

\textit{Limitations and outlook.} Judge-based metrics may reflect model-family bias and prompt sensitivity; while HH/PP controls provide useful context, they do not eliminate all sources of bias.
Our experiments primarily use a fixed assistant configuration and limited randomization, and our coverage is restricted to four English-centric datasets.
Additionally, goal synthesis relies on an auxiliary LLM that produces behavioral abstractions of user intent; while this avoids verbatim leakage, the summarization process may smooth over atypical or incoherent user behavior, potentially causing the benchmark to underrepresent edge-case human interaction patterns. Furthermore, the current protocol evaluates proxies against a fixed reference trajectory by conditioning the assistant on the reference history; evaluating proxies in fully open-ended rollouts, where the assistant has no access to the reference requires a different evaluation methodology.
Finally, lexical diversity captures surface-level distributional variation and does not fully represent interaction-level diversity such as repair, initiative, or long-horizon consistency.

Overall, \textsc{MirrorBench} is intended as a benchmark for systematically comparing user proxies and for grounding progress on realistic user simulation in conversational AI.

\bibliographystyle{ACM-Reference-Format}
\bibliography{bibfile}

%%
%% If your work has an appendix, this is the place to put it.
\appendix

\section{Extended Quantitative Results}
\label{app:extended_results}

\paragraph{\textbf{Human-Judge Correlation Across Datasets and Proxies.}}

Table~\ref{tab:human_judge_extended} extends the human-judge correlation analysis from \S\ref{subsec:results->human_judge_correlation} beyond the single (Gemini-2.5-Pro, ChatbotArena) condition reported in the main paper. We provided blinded human annotators to score 50-100 episodes per condition across three datasets (ChatbotArena, OASST1, ClariQ) and three proxy models (Gemini-2.5-Pro, Claude-4-Sonnet, GPT-4o), then computed Spearman's $\rho$ between human scores and Claude-4-Sonnet judge scores. GTEval $\rho$ ranges from 0.607 to 0.697 and PI $\rho$ from 0.532 to 0.671, all $p < 0.001$. The consistent alignment across diverse dataset domains and proxy models confirms that the judge reliability observed in the main paper is not specific to a single condition, and that Claude-4-Sonnet is a stable judge across the experimental configurations used in MirrorBench.

\begin{table}[h]
\centering
\small
\caption{Extended human-judge correlation. Spearman's $\rho$ between Claude-4-Sonnet judge scores and blinded human annotations across four (proxy, dataset) conditions; all $p < 0.001$.}
\label{tab:human_judge_extended}
\setlength{\tabcolsep}{4pt}
\begin{tabular}{llccc}
\toprule
Dataset & Proxy & $N$ & GTEval $\rho$ & PI $\rho$ \\
\midrule
ChatbotArena & Gemini-2.5-Pro  & 100 & 0.697 & 0.608 \\
OASST1       & Claude-4-Sonnet & 50  & 0.624 & 0.671 \\
OASST1       & GPT-4o          & 50  & 0.607 & 0.532 \\
ClariQ       & GPT-4o          & 50  & 0.686 & 0.596 \\
\bottomrule
\end{tabular}
\end{table}

\paragraph{\textbf{Multi-Seed Robustness Across All Datasets.}}

Table~\ref{tab:multiseed_extended} extends the multi-seed robustness analysis from \S\ref{subsec:results->multi_seed} beyond the 5-seed ChatbotArena runs shown in Figure~\ref{fig:multi_seed_robustness}. We ran 3 additional seeds on ClariQ, OASST1, and Qulac for three proxy models (GPT-4o, Claude-4-Sonnet, Gemini-2.5-Pro) with the assistant and judge fixed. GTEval mean\,$\pm$\,SD is reported for all 12 conditions. SD is $\leq$\,0.016 in all cases, confirming that the low run-to-run variance observed on ChatbotArena generalizes across datasets. Proxy rankings are stable in 11 of 12 conditions; the single exception is OASST1 where Gemini-2.5-Pro and GPT-4o differ by only 0.004 (effectively tied). The canonical ordering Gemini-2.5-Pro $>$ GPT-4o $>$ Claude-4-Sonnet holds across all four datasets.

\begin{table}[h]
\centering
\tiny
\caption{Multi-seed GTEval stability (mean\,$\pm$\,SD) across all four datasets. 5 seeds for ChatbotArena, 3 for the remaining datasets. SD\,$\leq$\,0.016 in all 12 conditions.}
\label{tab:multiseed_extended}
\setlength{\tabcolsep}{4pt}
\begin{tabular}{lcccc}
\toprule
Proxy & ChatbotArena & ClariQ & OASST1 & Qulac \\
\midrule
GPT-4o          & $0.554 \pm 0.000$ & $0.239 \pm 0.002$ & $0.614 \pm 0.000$ & $0.351 \pm 0.001$ \\
Claude-4-Sonnet & $0.445 \pm 0.000$ & $0.210 \pm 0.002$ & $0.490 \pm 0.007$ & $0.288 \pm 0.000$ \\
Gemini-2.5-Pro  & $0.610 \pm 0.000$ & $0.296 \pm 0.007$ & $0.618 \pm 0.016$ & $0.502 \pm 0.013$ \\
\bottomrule
\end{tabular}
\end{table}

\section{Goal-Conditioning Ablation: Qualitative Examples}
\label{app:goal_ablation_examples}

Table~\ref{tab:goal_ablation_ex_chatbot} illustrates the qualitative difference between \texttt{goal\_full} and \texttt{goal\_topic} conditioning on two representative ChatbotArena episodes. In each case, both conditions produce fluent, human-sounding turns (which is why PI stays high under \texttt{goal\_topic}), but \texttt{goal\_topic} causes the proxy to pursue a plausible \emph{related} question rather than the user's actual intent (which is why GTEval drops).

\begin{table*}[h]
\centering
\small
\caption{Qualitative comparison of \texttt{goal\_full} vs.\ \texttt{goal\_topic} conditioning on two ChatbotArena episodes (GPT-4o proxy). \texttt{goal\_full} stays on the user's specific request; \texttt{goal\_topic} deflects to a generic related question.}
\label{tab:goal_ablation_ex_chatbot}
\setlength{\tabcolsep}{5pt}
\begin{tabular}{p{0.10\linewidth} p{0.27\linewidth} p{0.27\linewidth} p{0.27\linewidth}}
\toprule
\textbf{Episode} & \textbf{Real user turn} & \textbf{Proxy turn (\texttt{goal\_full})} & \textbf{Proxy turn (\texttt{goal\_topic})} \\
\midrule
JavaScript ArrayBuffer
&
\textit{``Please write a JavaScript function that receives an ArrayBuffer, parses it and returns it as a JavaScript object.''}
&
``Can you include an example of how to \textbf{use the function} with a sample ArrayBuffer?''
&
``Can you explain how to \textbf{convert an ArrayBuffer to a string} in JavaScript?''
\\[6pt]
Speed comparison
&
\textit{``mark is faster than mary. mary is faster than joe. is joe faster than mark?''}
&
``If Mark is faster than Mary, and Mary is faster than Joe, does that mean \textbf{Mark is also faster than Joe}?''
&
``Who is faster, \textbf{Usain Bolt or a cheetah}?''
\\
\bottomrule
\end{tabular}
\end{table*}

In both examples, the \texttt{goal\_full} proxy pursues the user's concrete objective (e.g., asking for a code usage example, staying on the same logical reasoning problem), while \texttt{goal\_topic} produces a related but \emph{generic} follow-up that a human asking the original question would not ask. This drift is captured by the drop in GTEval under \texttt{goal\_topic} despite high PI, validating the interpretation in \S\ref{subsec:goal_ablation}.

\section{MirrorBench System}
\label{appsec:system}

In this section, we explain the system aspects of the benchmarking framework. Although the metric definitions and execution may seem straightforward, substantial work went into creating a scalable, extensible system for flexible research use. We provide a detailed architectural blueprint in \S\ref{appsubsec:system->arch}, explain the execution flow in \S\ref{appsubsec:system->exec}, and provide metric implementation details in \S\ref{appsubsec:system->metric_impl}

\subsection{Architecture}
\label{appsubsec:system->arch}

\begin{figure*}[t]
      \centering
      \includegraphics[width=\textwidth,
        trim={0mm 0mm 0mm 0mm}, % L B R T — tweak these numbers
        clip,
        keepaspectratio]{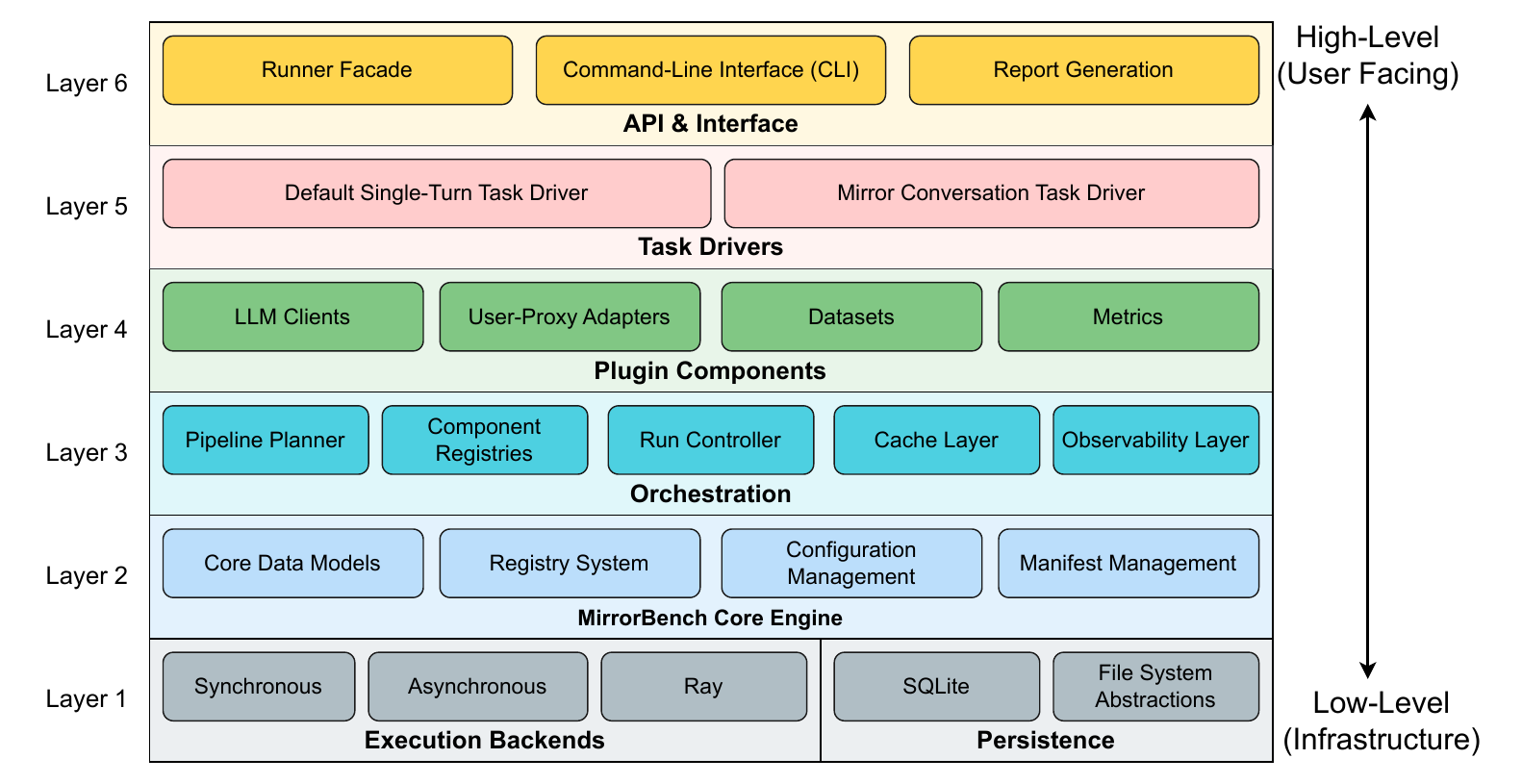}
      \caption{MirrorBench Architecture: Six-layer stack from low-level execution backends \& persistence up through the core engine, plugin components, CLI \& reporting, and task drivers. Top layers are user-facing; bottom layers are low-level infrastructure abstractions.}
      \label{fig:architecture_layers}
\end{figure*}

MirrorBench is a six-layer stack that cleanly separates infrastructure from evaluation logic (Fig.~\ref{fig:architecture_layers}). Layers build upward, lower layers provide execution and data services; upper layers host pluggable components, task logic, and user interfaces.

\subsubsection{Execution Backends \& Persistence}
\label{subsubsec:framework->overview->layer1}

This layer is responsible for all the execution and data management. The layer consists of two high-level modules explained below.

\paragraph{\textbf{Execution Backend}} It is responsible for taking decomposed units of tasks as input and send them for execution on the respective execution engine. The framework provides abstractions to support multiple execution backends to execute evaluation jobs at scale. The abstract interface provides uniform schema for implementations to extend the support for new execution engines. The framework supports fully-functional synchronous \& asynchronous execution backends, and provides stub implementation to extend for Ray \cite{10.5555/3291168.3291210} backend.

\paragraph{\textbf{Persistence}} All evaluation artifacts are persisted in a SQLite database organized around a three-level hierarchy: \textit{runs} define experiment configurations, \textit{units} specify individual evaluation tasks (parameterized by user-proxy, dataset, metric, and seed), and \textit{episodes} capture per-trial results including execution duration, metric values, and telemetry. Aggregate statistics (mean, standard deviation, confidence intervals) are materialized in a separate \texttt{metrics} table to support efficient querying, while composite scores across multiple metrics are stored in \texttt{scorecards} for holistic comparisons. The complete schema is detailed in Table~\ref{tab:db-schema}.

\begin{table*}[!tbp]
  \centering
  \caption{Complete MirrorBench Run Database Schema}
  \label{tab:db-schema}
  \small
  \begin{tabular}{@{}lllp{6.5cm}@{}}
  \toprule
  \textbf{Table} & \textbf{Field} & \textbf{Type} & \textbf{Description} \\
  \midrule
  \multirow{7}{*}{\texttt{runs}}
    & run\_id & TEXT & Unique identifier for the experiment run (primary key) \\
    & created\_at & TEXT & ISO-8601 timestamp of run creation \\
    & status & TEXT & Run status: \texttt{created}, \texttt{running}, \texttt{completed}, \texttt{failed} \\
    & engine & TEXT & Execution engine identifier (e.g., \texttt{sync}, \texttt{async}) \\
    & planner\_version & TEXT & Version of the planner used for the run \\
    & summary\_json & TEXT & JSON-encoded run-level summary statistics \\
    & notes & TEXT & User-provided annotations or metadata \\
  \midrule
  \multirow{8}{*}{\texttt{units}}
    & run\_id & TEXT & Foreign key to parent run \\
    & unit\_id & TEXT & Unique identifier for the evaluation unit (composite primary key) \\
    & user\_proxy & TEXT & User-proxy agent implementation identifier \\
    & dataset & TEXT & Dataset name (e.g., \texttt{oasst1}, \texttt{clariq}) \\
    & metric & TEXT & Evaluation metric (e.g., \texttt{gteval}) \\
    & seed & INTEGER & Random seed for reproducibility \\
    & judge & TEXT & Optional judge model for LLM-based metrics \\
    & status & TEXT & Unit execution status: \texttt{pending}, \texttt{running}, \texttt{completed} \\
  \midrule
  \multirow{8}{*}{\texttt{episodes}}
    & run\_id & TEXT & Foreign key to parent run \\
    & unit\_id & TEXT & Foreign key to parent unit \\
    & episode\_id & TEXT & Unique episode identifier (primary key) \\
    & status & TEXT & Episode outcome: \texttt{success}, \texttt{failure}, \texttt{timeout} \\
    & duration\_s & REAL & Wall-clock execution time in seconds \\
    & artifact\_path & TEXT & File path to serialized conversation artifact \\
    & summary & TEXT & Human-readable episode summary \\
    & metric\_values & TEXT & JSON-encoded per-metric scores for this episode \\
    & telemetry\_json & TEXT & JSON-encoded execution telemetry (token counts, API calls) \\
  \midrule
  \multirow{8}{*}{\texttt{metrics}}
    & run\_id & TEXT & Foreign key to parent run \\
    & unit\_id & TEXT & Foreign key to parent unit \\
    & metric & TEXT & Registered metric name (primary key) \\
    & mean & REAL & Sample mean across episodes \\
    & standard\_deviation & REAL & Sample standard deviation \\
    & confidence\_interval & REAL & 95\% confidence interval half-width \\
    & p\_value & REAL & Statistical significance (e.g., vs.\ baseline) \\
    & sample\_size & INTEGER & Number of valid episodes \\
    & extras & TEXT & JSON-encoded additional statistics \\
  \midrule
  \multirow{4}{*}{\texttt{telemetry}}
    & run\_id & TEXT & Foreign key to parent run \\
    & unit\_id & TEXT & Foreign key to parent unit \\
    & key & TEXT & Telemetry key (e.g., \texttt{total\_tokens}, \texttt{api\_errors}) \\
    & value & TEXT & Aggregated telemetry value (JSON or scalar) \\
  \midrule
  \multirow{6}{*}{\texttt{scorecards}}
    & run\_id & TEXT & Foreign key to parent run \\
    & name & TEXT & Scorecard identifier (e.g., \texttt{realism}, \texttt{diversity}) \\
    & score & REAL & Weighted composite score \\
    & weights & TEXT & JSON-encoded metric weights used for aggregation \\
    & missing\_metrics & TEXT & JSON list of metrics excluded due to missing data \\
    & extras & TEXT & JSON-encoded additional scorecard metadata \\
  \bottomrule
  \end{tabular}
\end{table*}

\subsubsection{Core Engine}
\label{subsubsec:framework->overview->layer2}

All the core functionalities like data models, registry builder, configuration and manifest management are contained in Layer 2 of the architecture. Since Layer 1 provides uniform abstractions to produce and consume objects, this layer considers backend-agnostic implementation.

\paragraph{\textbf{Data Models}} All the data structures for messages, episodes, evaluation units and run manifests are defined with typed data models using \texttt{TypedDict} and \texttt{pydantic}. We formally explain the context of these different data objects later.

\paragraph{\textbf{Registry Builder}} The framework provides certain high-level components which requires module-specific registries. This layer provides the abstract registry builder that supports creating metadata-aware registry for the downstream modules. Such registries can avoid silent errors since we can ensure certain constraints on the classes being registered with the metadata.

\paragraph{\textbf{Configuration \& Manifest Management}} A typical evaluation job requires configuring user-proxy, datasets, metrics, scorecards etc. We define standard configuration schema for each of such components. Moreover, upon invocation of job or during dryrun, the system decomposes evaluation task into small evaluation units. In order to support parallelization (for async and remote backends), we persist self-contained information about the executable units as manifest files. 

\subsubsection{Orchestration}
\label{subsubsec:framework->overview->layer3}

This layer coordinates the evaluation workflow by managing component lifecycles, execution planning, and runtime operations. The layer consists of five orchestration modules.

\paragraph{\textbf{Pipeline Planner}} It validates job configurations and generates execution plans by decomposing evaluation jobs into granular evaluation units (user-proxy, dataset, metric, seed combinations). The planner performs compatibility filtering based on component metadata to ensure valid combinations and produces reproducible plan manifests.

\paragraph{\textbf{Component Registries}} It maintains metadata-aware registries for all pluggable components including user-proxy adapters, datasets, metrics, model clients, and judges. Registries support dynamic component discovery and instantiation through decorator-based registration patterns.

\paragraph{\textbf{Run Controller}} It orchestrates end-to-end evaluation runs by managing execution flow, coordinating backend selection (synchronous, asynchronous, or Ray), tracking progress, and handling run lifecycle including resumption of interrupted runs.

\paragraph{\textbf{Cache Layer}} It provides SQLite-based caching for model responses to reduce redundant API calls and improve evaluation efficiency. The cache uses content-based keys (hashing model name, messages, and configuration) and supports namespace isolation and TTL-based expiration.

\paragraph{\textbf{Observability Layer}} It provides structured logging via \texttt{structlog} and optional OpenTelemetry integration for distributed tracing and metrics collection. The layer supports multiple output formats (JSON, text) and configurable log levels for debugging and monitoring.

\subsubsection{Plugin Components}
\label{subsubsec:framework->overview->layer4}

This layer contains the core pluggable components that implement domain-specific evaluation logic. All components follow standardized interfaces and register through the registry system.

\paragraph{\textbf{Model Clients}} It wraps provider SDKs (OpenAI, LangChain, Azure, etc.) with a unified interface for LLM invocation. Clients automatically capture telemetry (tokens, latency, cost) and support transparent caching through wrapper composition.

\paragraph{\textbf{User-Proxy Adapters}} It adapts diverse agent frameworks and LLM configurations into a standardized \texttt{AgentSession} interface. Adapters normalize differences in message formats, tool usage patterns, and execution models across frameworks.

\paragraph{\textbf{Datasets}} It loads and normalizes evaluation datasets from various sources (HuggingFace, JSONL files, etc.) into standardized episode objects. Loaders support dataset splitting, sampling limits, and provide metadata about available splits and reference statistics.

\paragraph{\textbf{Metrics}} It implements human-likeness measurements including lexical diversity metrics (\textsc{MATTR}, \textsc{HD-D}, \textsc{Yule's~K}) and LLM-as-judge metrics (\textsc{GTEval}, \textsc{PI}, \textsc{RNR}). Metrics declare compatibility requirements through metadata (task types, judge dependencies, reference statistics needs).

\subsubsection{Task Drivers}
\label{subsubsec:framework->overview->layer5}

This layer implements interaction protocols that orchestrate how the artificial conversation between user-proxy \& assistant agent is synthesized and how it is engaged with the evaluation metrics. Task drivers bridge the gap between high-level evaluation specifications and low-level execution.

\paragraph{\textbf{Default Single-Turn Task Driver}} It executes single-turn interactions where the user-proxy receives episode context and generates a single response. This driver is suitable for simple prompt-completion tasks and turn-level utterance evaluation.

\paragraph{\textbf{Mirror Conversation Task Driver}} It orchestrates multi-turn conversational interactions by executing sequences of user-proxy responses within episodic contexts. This driver supports conversation-level evaluations where human-likeness is assessed across extended interactions.

\subsubsection{API \& Interface}
\label{subsubsec:framework->overview->layer6}

This layer provides user-facing interfaces for interacting with the framework through programmatic APIs and command-line tools.

\paragraph{\textbf{Runner Facade}} It provides a simplified programmatic API for executing evaluation runs from Python code. The facade abstracts orchestration complexity and exposes high-level methods for planning, execution, and result retrieval with progress tracking.

\paragraph{\textbf{Command-Line Interface (CLI)}} It exposes framework functionality through a comprehensive CLI supporting workflow commands (\texttt{plan}, \texttt{dryrun}, \texttt{run}), reporting (\texttt{report json}), run management (\texttt{runs inspect}, \texttt{runs delete}), and cache operations (\texttt{cache stats}, \texttt{cache purge}).

\paragraph{\textbf{Report Generation}} It produces structured evaluation summaries with aggregated statistics including means, confidence intervals, and significance testing. The module supports JSON report output with planned support for HTML visualization.

\subsection{Execution Flow}
\label{appsubsec:system->exec}

\begin{figure}[t]
    \centering
    \includegraphics[
        width=0.7\columnwidth,
        trim={0mm 19mm 0mm 20mm}, % L B R T — tweak these numbers
        clip,
        keepaspectratio
    ]{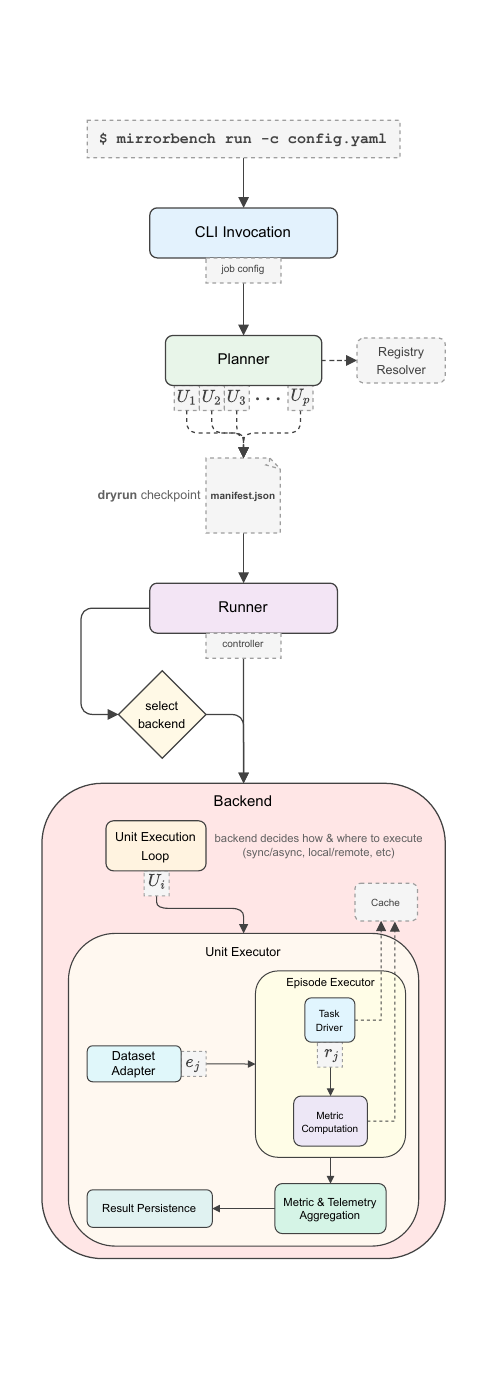}
    \caption{MirrorBench execution flow. The framework decomposes an evaluation job into \emph{units} $\{U_1,\ldots,U_p\}$; each unit \(U_i\) iterates over \emph{episodes} $\{e_1,\ldots,e_n\}$ produced by the dataset adapter and executed via task drivers. Metrics are computed per episode and aggregated within each unit with confidence intervals.}
    \label{fig:execution_flow}
\end{figure}

This section details the end-to-end execution that instantiates the pipeline $\Psi_T$ (as described in §\ref{subsec:mirrorbench->benchmark_proto}).  Figure~\ref{fig:execution_flow} shows the path from CLI invocation through planning, orchestration, and nested execution to persisted results, compiling high-level specifications into reproducible evaluation artifacts.

\paragraph{\textbf{CLI Invocation and Planning}}
An evaluation job is declared via a configuration file (appendix \S\ref{app:config_manifest_report}) that specifies user proxies \(\mathcal{P}=\{\theta_{u_1},\ldots,\theta_{u_q}\}\), datasets \(\mathcal{D}=\{D_1,\ldots,D_d\}\), the metric set \(M=\{m_1,\ldots,m_k\}\), a fixed assistant model \(\theta_a\), and run parameters (backend, observability, maximum concurrency, and a seed set \(S=\{s_1,\ldots,s_r\}\)). The seed set enables repeated trials of the same combination under distinct random initializations, supporting variance estimation and confidence intervals.

The \emph{Planner} validates the configuration against component registries and enumerates all mutually compatible tuples to produce execution units \(U=\{U_1,\ldots,U_p\}\).
\[
    U=\left\{
        (\theta_{u_x}, D_y, m_z, s_w)\,
        \middle|\,
    \begin{aligned}
        &\theta_{u_x}\in \mathcal{P},\; D_y\in\mathcal{D},\\
        &m_z\in M,\; s_w\in S,\\
        &\operatorname{Compat}(\theta_{u_x}, D_y, m_z)
    \end{aligned}
    \right\}
\]
Here \(\operatorname{Compat}(\cdot)\) returns \texttt{true} iff the components can be jointly executed under their declared capabilities and requirements (e.g. metric prerequisites, model/back-end constraints). Each unit \(U_i=(\theta_{u_x}, D_y, m_z, s_w)\) is executed and evaluated independently, with models, prompts, and driver assignments fixed at plan time. The planner creates \& persists a replayable \emph{manifest} file (appendix \S\ref{app:config_manifest_report}) capturing all requisite metadata, and exposes a \texttt{dryrun} checkpoint for inspection prior to committing computational resources.

% \[
%     U=\left\{
%         (\theta_{u_x}, D_y, m_z, \theta_a, s_w)\,
%         \middle|\,
%     \begin{aligned}
%         &\theta_{u_x}\in \mathcal{P},\; D_y\in\mathcal{S},\; m_z\in M,\; s_w\in S,\\
%         &\operatorname{Compat}(\theta_{u_x}, D_y, m_z, \theta_a)
%     \end{aligned}
%     \right\}.
% \]

\paragraph{\textbf{Orchestration and Backend Selection}}
The \emph{Runner} instantiates a \emph{Run Controller} that manages execution state, telemetry, and result persistence. The controller initializes a run-specific SQLite database to store task driver outputs and metric values. Based on the run configuration, the system selects an available backend (sync, async, or distributed) and dispatches the list of units $U$ for execution under the chosen strategy. For each unit $U_i \in U$, the backend invokes a unit executor that loads the dataset, instantiates the proxy, initializes the metrics, and configures the task driver.

\paragraph{\textbf{Unit Execution and Episode Extraction}}
For each unit \(U_i = (\theta_{u_x}, D_y, m_z, s_w)\), the unit executor loads dataset \(D_y\) and enumerates its constituent episodes. Each episode \(e_j\) is an immutable specification of a single reference dialogue (§\ref{subsec:mirrorbench->benchmark_proto}), with dataset layout and fields given by
\[
\begin{aligned}
D_y \;&=\; \{\, e_1, e_2, \ldots, e_{n} \,\},\\[2pt]
e_j \;&=\; \bigl(h_j^{\mathrm{ref}},\, g_j,\, \mathrm{meta}_j\bigr),\\[2pt]
h_j^{\mathrm{ref}} \;&=\; \bigl[(u_{j,1}^{\mathrm{ref}}, a_{j,1}^{\mathrm{ref}}), \ldots, (u_{j,L_j}^{\mathrm{ref}}, a_{j,L_j}^{\mathrm{ref}})\bigr],
\end{aligned} 
\]
where \(g_j\) denotes the user goal (explicit or inferred), and \(\mathrm{meta}_j\) is an optional field containing evaluation-specific annotations (e.g., task tags or reference statistics). The unit executor iterates over the episode set \(\{e_j\}\), delegating each episode to the appropriate episode executor for execution. Episode executor first synthesizes proxy-assistant conversation using task driver and later compute evaluation results using metric implementation \(m_z\).

\paragraph{\textbf{Conversation Synthesis via Task Drivers}}
For each episode \(e_j\), the task driver orchestrates the synthesis of a proxy-assistant interaction by alternating invocations of the user proxy \(\theta_{u_x}\) (abbrev.\ \(\theta_u\)) and the assistant model \(\theta_a\), matching the reference dialogue length \(L_j = |h_j^{\mathrm{ref}}|\).  Let \(\hat{h}_{j,t}\) denote the synthetic history after \(t\) turns, with \(\hat{h}_{j,0}=\langle\,\rangle\). The synthetic rollout for episode \(e_j\) is:
\[
\begin{aligned}
&\textbf{Initialize:}\quad \hat{h}_{j,0} \gets \langle\,\rangle \\[2pt]
&\textbf{For } t = 1,\ldots,L_j \textbf{:}\\
&\qquad \hat{u}_{j,t} \;\gets\; \theta_u\!\bigl(\hat{h}_{j,t-1},\, g_j\bigr), \\[2pt]
&\qquad \hat{a}_{j,t} \;\gets\; \theta_a\!\bigl(\hat{h}_{j,t-1} \mathbin{\|} (\hat{u}_{j,t}), h_j^{\mathrm{ref}}\bigr), \\[2pt]
&\qquad \hat{h}_{j,t} \;\gets\; \hat{h}_{j,t-1} \mathbin{\|} (\hat{u}_{j,t}, \hat{a}_{j,t}).
\end{aligned}
\]
% \[
% \begin{aligned}
% \text{for } t &= 1 \text{ to } T_j: \\
% u_{j,t}^{\mathrm{syn}} &= \theta_u(h_j^{\mathrm{syn}}, g_j), \\[2pt]
% a_{j,t}^{\mathrm{syn}} &= \theta_a(h_j^{\mathrm{syn}} \cup \{u_{j,t}^{\mathrm{syn}}\}), \\[2pt]
% h_j^{\mathrm{syn}} &\gets h_j^{\mathrm{syn}} \cup \{(u_{j,t}^{\mathrm{syn}}, a_{j,t}^{\mathrm{syn}})\}.
% \end{aligned}
% \]
The driver records per-turn telemetry (latency and token usage) and produces the \emph{episode artifact}
\[
r_j \;=\; \bigl(\,\hat{h}_{j,L_j},\; e_j,\; \mathrm{telemetry}_j\bigr),
\]
which binds the synthetic transcript to its reference. The synthetic history is persisted in cache, ensuring that other metric computations can reuse it.

\begin{table*}[!tbp]
  \centering
  \caption{Experimental Configuration and Dependency Versions}
  \label{tab:experimental-setup}
  \small
  \begin{tabular}{@{}llp{7cm}@{}}
  \toprule
  \textbf{Category} & \textbf{Component} & \textbf{Version/Configuration} \\
  \midrule
  \multicolumn{3}{l}{\textit{Hardware \& Infrastructure}} \\
  \midrule
  OS & Linux & SUSE Linux Enterprise Server 15 SP4 (kernel 5.14+) \\
  CPU & Standard & Intel Xeon Platinum 8468 (16+ cores) \\
  RAM & Standard & 64 GB DDR4 \\
  GPU & GPT-OSS-120B only & 4$\times$ NVIDIA H200 (140GB VRAM each) or equivalent \\
  CUDA & GPT-OSS-120B only & CUDA 12.4+ \\
  \midrule
  \multicolumn{3}{l}{\textit{Runtime \& Storage}} \\
  \midrule
  Python & Interpreter & 3.12.0+ \\
  SQLite & Database & 3.45.0+ (WAL mode enabled) \\
  \midrule
  \multicolumn{3}{l}{\textit{API Providers (accessed Oct--Dec 2024)}} \\
  \midrule
  OpenAI & GPT-4o & \texttt{gpt-4o-2024-08-06
} \\
   & GPT-5 & \texttt{gpt-5-2025-08-07} \\
  Anthropic & Claude-4-Sonnet & \texttt{claude-sonnet-4-20250514} \\
  Vertex AI & Gemini-2.5-Pro & \texttt{2025-06-17} \\
  Self-hosted & GPT-OSS-120B & vLLM 0.10.1+gptoss, BF16 precision \\
  \midrule
  \multicolumn{3}{l}{\textit{Core Library Versions}} \\
  \midrule
  Model Clients & \texttt{openai} & 1.40.0+ \\
   & \texttt{langchain-openai} & 0.1.0+ \\
  Data & \texttt{pydantic} & 2.5.0+ \\
   & \texttt{datasets} & 4.1.1+ (HuggingFace) \\
  Utilities & \texttt{tiktoken} & 0.7.0+ (tokenization) \\
   & \texttt{nltk} & 3.9.2+ (sentence segmentation) \\
   & \texttt{tenacity} & 8.2.3+ (retry logic) \\
  Logging & \texttt{structlog} & 24.1.0+ \\
  Telemetry & \texttt{opentelemetry-sdk} & 1.37.0+ \\
   & \texttt{opentelemetry-exporter-otlp} & 1.37.0+ \\
  \bottomrule
  \end{tabular}
\end{table*}

\paragraph{\textbf{Metric Computation}}
After producing an episode artifact \(r_j\), metric \(m_z\) evaluates \(r_j\) to yield a scalar value \(v_{j,m_z}\). Depending on the metric type (e.g., lexical, or LLM-judge), the computation may involve reference comparison, pairwise scoring, or external model inference. Formally,
\[
v_{j,m_z} \;=\; m_z\!\bigl(r_j\bigr),
\]
where \(v_{j,m_z}\in\mathbb{R}\) represents the realism score for metric \(m_z\) on episode \(e_j\).  Each evaluation may also produce auxiliary data such as judge verdicts, raw reasoning traces, or calibration diagnostics. The full episode artifact, together with all metric outputs, is persisted to disk for aggregation, post-hoc inspection and reproducibility.

\paragraph{\textbf{Result Aggregation}}
After processing all episodes \(\{e_1,\ldots,e_{n}\}\) within unit \(U_i\), the executor aggregates the metric outputs into summary statistics.  For metric \(m_z\), let \(V_{U_i,m_z} = \{v_{1,m_z}, v_{2,m_z}, \ldots, v_{n,m_z}\}\) denote the set of episode-level scores. The unit-level aggregate includes the mean, standard deviation, and \(95\%\) confidence interval:
\begin{align*}
\bar{v}_{U_i,m_z} \;&=\; \frac{1}{n} \sum_{j=1}^{n} v_{j,m_z}, \\ 
s_{U_i,m_z} \;&=\; 
\sqrt{\frac{1}{n}\sum_{j=1}^{n}(v_{j,m_z}-\bar{v}_{U_i,m_z})^2}, \\
\mathrm{CI}_{95}(U_i,m_z)
&=
\left[
\begin{array}{@{}l@{}}
\bar{v}_{U_i,m_z} - 1.96\,\dfrac{s_{U_i,m_z}}{\sqrt{n}} \\[6pt]
\bar{v}_{U_i,m_z} + 1.96\,\dfrac{s_{U_i,m_z}}{\sqrt{n}}
\end{array}
\right].
\end{align*}
The run database stores these aggregates, telemetry totals, and per-episode status indicators (success/failure).

\paragraph{\textbf{Iteration and Reporting}}
The backend executes each unit \(U_i=(\theta_{u_x},D_y,m_z,s_w)\) in \(U=\{U_1,\ldots,U_p\}\). Upon completion, the controller updates the last-run pointer for convenient access and consolidates a unified \texttt{RunSummary} that collates, for every \(m_z\) and unit \(U_i\), the tuple \(\bigl(\bar{v}_{U_i,m_z},\,s_{U_i,m_z},\,\mathrm{CI}_{95}(U_i,m_z)\bigr)\), along with cumulative telemetry and execution metadata.  This summary serves as the canonical artifact for subsequent JSON/HTML report generation (appendix \S\ref{app:config_manifest_report}) and cross-proxy comparison. The caching layer eliminates repeated LLM calls across units/runs, and the plan manifest enables exact replays of component instantiation and invocation for reproducible results.

\subsection{Implementation Details of Metrics}
\label{appsubsec:system->metric_impl}

Below, we provide the implementation details for the metrics and other associated components in the evaluation pipeline. Additionally, in Table \ref{tab:experimental-setup}, we provide details of the machine we used to run all the experiments mentioned in this paper.

\paragraph{\textbf{Tokenization}} All lexical metrics use the GPT-4o tokenizer (via \texttt{tiktoken}) to ensure consistency with modern LLM vocabularies. User utterances in the dialogue sample are concatenated with space separators before tokenization.

\paragraph{\textbf{Judge Caching}} Judge responses are cached using content-based keys (hash of model name, messages, temperature, and metric parameters) with namespace isolation per metric. Default TTL is 30 days. This enables reproducible experiments and cost reduction in iterative evaluation.

\paragraph{\textbf{Multiple Judge Calls}} By default, each judge metric runs $c$ independent samples per episode ($c$ is a metric parameter to be set) with different random seeds (for PI's order randomization). Final scores are averaged; individual samples and reasoning traces are stored in metadata for audit.

\paragraph{\textbf{Statistical Reporting}} Aggregated metrics report mean, standard deviation, and 95\% confidence intervals using Student's $t$-distribution (Equation \ref{eq:ci}). The framework optionally supports bootstrap confidence intervals (1000 iterations) for non-parametric estimation.

\paragraph{\textbf{Aggregation Across Episodes}} For each metric $M$ and proxy-dataset pair, the framework computes per-episode scores $\{s_1, s_2, \ldots, s_n\}$ and aggregates them using mean, standard deviation, and confidence intervals. For lexical metrics, z-scores are computed per-episode using the human baseline statistics, then aggregated. For judge metrics, raw scores are averaged first; if calibration is enabled, HH and PP statistics are computed separately and applied to normalize the aggregated results.

\paragraph{\textbf{Handling Missing References}} Episodes missing human reference conversations (required for GTEval, PI, and z-score computation) are excluded from metric computation. The framework logs these exclusions and reports the count of valid episodes in the metadata. This ensures that all reported statistics are based on complete, valid comparisons.

\paragraph{\textbf{Stability Filters}} Lexical metrics require a minimum token count (default: 5) to produce stable estimates. Episodes below this threshold are flagged as unstable and optionally excluded from aggregation. This prevents outlier scores from very short utterances (e.g., single-word responses) from skewing results.

\paragraph{\textbf{LLM Settings}} All the LLMs used during the evaluation have fixed \texttt{temperature=0} and \texttt{max\_tokens=2048}. Moreover, all the LLM clients have by default \emph{``retry with exponential backoff"} scheme enabled where \texttt{backoff\_duration=2s} and \texttt{max\_retries=5}.

\paragraph{\textbf{Telemetry \& Observability}} The framework captures per-invocation metrics (latency-to-first-token, generation time, token counts, cost) and organizes traces into hierarchical spans (\texttt{run}, \texttt{unit}, \texttt{episode}) annotated with run/unit identifiers, status, and dataset metadata. Structured logs use \texttt{structlog} with JSON formatting and contextual bindings.

\paragraph{\textbf{Model Pricing}.} Cost estimates reported in our experiments are computed using the provider-published pricing as of 25 October 2025, shown in Table~\ref{tab:model-pricing}. All costs are calculated in USD per million tokens for both input (prompt) and output (completion) tokens. Token counts are obtained directly from provider API responses. The total cost $\text{C}_{\text{USD}}$ for each episode is computed as

\[
\text{C}_{\text{USD}} =
   \frac{\text{tokens}_{\text{input}} \times \text{price}_{\text{input}} + \text{tokens}_{\text{output}} \times
  \text{price}_{\text{output}}}{10^6}
\]

Later, we aggregate across all invocations within a unit. Note that pricing may vary by region, deployment tier, or provider discounts; we report costs using standard public API rates.

\begin{table}[t]
  \centering
  \caption{Model Pricing (USD per million tokens). Pricing as of 25 October 2025 from the provider's public APIs.}
  \label{tab:model-pricing}
  \small
  \begin{tabular}{@{}lrr@{}}
  \toprule
  \textbf{Model} & \textbf{Input (prompt)} & \textbf{Output (completion)} \\
  \midrule
  GPT-4o & \$2.50 & \$10.00 \\
  GPT-5 & \$1.25 & \$10.00 \\
  Claude 4 Sonnet & \$3.00 & \$15.00 \\
  Gemini 2.5 Pro & \$1.25 & \$10.00 \\
  \bottomrule
  \end{tabular}
\end{table}

\section{Prompts, Config, Manifest \& Report}
\label{app:prompts-configs}

This appendix provides the exact artifacts required to reproduce our runs: judge prompts, task-driver prompts, job configuration, and the emitted plan manifest. For reproducibility, we report the prompt, model identifiers, and key decoding parameters (temperature, max tokens).

\subsection{System Prompts}
\label{app:system-prompts}

We provide system prompts used across the evaluation pipeline (metric judges, user-proxy, assistant).

\begin{tcolorbox}[enhanced jigsaw,breakable,pad at break*=1mm,
  colback=blue!5!white,colframe=blue!75!black,title=PI Judge Prompt]
[System]  

Please act as an impartial judge and choose the most realistic USER among the two conversations provided below. You should choose the conversation which has USER turns written by a real user in your opinion. Based on your experience \& knowledge of human conversation, determine which USER in which conversation sounds more like a real human interacting with a chatbot. Note that your judgement should be based on style, tone and behavior of USER rather than the response quality of their utterances. Avoid any position biases and ensure that the order in which the responses were presented does not influence your decision. Do not allow the length of the USER messages to influence your evaluation. Do not let ASSISTANT messages influence your decision. Your evaluation must only be dependent on USER messages only. Be as objective as possible.  \\

Before giving your final evaluation verdict, you should first think through and produce your reasoning step-by-step. Then, based on your reasoning, give your final verdict on which conversation has USER turns that sound more like a real human interacting with a chatbot.  \\

If Conversation A is more real, then verdict would be ``A". If Conversation B is more real, then verdict would be ``B". If both are equally real or non-real, then verdict would be ``Tie".  \\

Your output must be of following JSON format: \{``reasoning": ``$<$short reason$>$", ``verdict": ``A"$|$``B"$|$``Tie"\}.  \\

[Conversation A]  

\{conversation\_a\}  \\

[Conversation B]  

\{conversation\_b\}  \\

Output ONLY valid JSON, no additional text.
\end{tcolorbox}
\noindent \begin{minipage}{\columnwidth}
\captionof{figure}{Judge prompt for Pairwise Indistinguishability Metric}
\label{fig_prompts:judge_pi}
\end{minipage}

\begin{tcolorbox}[enhanced jigsaw,breakable,pad at break*=1mm,
  colback=blue!5!white,colframe=blue!75!black,title=GTEval Judge Prompt]
You are an expert evaluator of conversational AI systems. Your task is to compare two conversations and assess whether the user responses in the proxy conversation exhibit similar style, tone, and realism to the user responses in the real conversation. \\

\#\# Real Conversation:  

\{real\_conversation\}  \\

\#\# Proxy Conversation (where user turns are generated by user proxy agent):  

\{proxy\_conversation\}  \\

\#\# Evaluation Criteria:  

1. **Style Similarity**: Do the proxy user responses match the conversational style of real user responses (formality, tone, verbosity)?  

2. **Realism**: Do the proxy user responses sound natural and human-like?  

3. **Contextual Appropriateness**: Are the proxy user responses appropriate given the conversation context? Note: You should not evaluate based on the content of the responses, only their style, realism, contextual appropriateness, and tone.  \\

\#\# Instructions:  

- Focus exclusively on comparing USER responses (ignore assistant responses)  

- Consider the overall conversation flow, not just individual turns  

- Evaluate style, tone, and naturalness without encouraging unsafe behaviour  

- If the comparison cannot be performed safely, explain the reason and provide a score of 0.0  

- Otherwise, provide detailed reasoning about the comparison and assign a score from 0.0 to 1.0 where:  

  - 0.0 = Proxy user responses are completely unlike real user responses  
  
  - 0.5 = Proxy user responses are somewhat similar but have noticeable differences  
  
  - 1.0 = Proxy user responses are highly similar in style and realism  \\

Provide your evaluation as a JSON object with the following structure:  

\{
    ``reasoning": ``Detailed explanation of your evaluation (2-3 sentences)",
    ``score": $<$float between 0.0 and 1.0$>$
\}  \\

Output ONLY valid JSON, no additional text.
\end{tcolorbox}
\noindent \begin{minipage}{\columnwidth}
\captionof{figure}{Judge prompt for GTEval Metric}
\label{fig_prompts:judge_gteval}
\end{minipage}

\begin{tcolorbox}[enhanced jigsaw,breakable,pad at break*=1mm,
  colback=blue!5!white,colframe=blue!75!black,title=RNR Judge Prompt]
[System]  

Please act as an impartial judge and provide verdict on the realism of the USER turns in the conversation provided below based on the provided rubric. You should avoid checking the responses of the ASSISTANT messages since they are not relevant to the task. Your evaluation should follow the rubric provided. Note that your judgement should be based on style, tone and behavior of USER rather than the response quality of their utterances.  \\

[Rubric for scoring USER realism]  

Provide verdict on how human-like the USER is on (``NO" or ``YES") using this rubric (verdict ``NO" means not real, verdict ``YES" means real):  

1. Concise and real-user like language  

2. Does not sound scripted or artificial  

3. Real-user like tone and style  \\

Return JSON: \{``reasoning": ``$<$1-2 sentences$>$", ``verdict": $<$``NO" or ``YES">\}.  \\

[Conversation]  

\{conversation\}  \\

Output ONLY valid JSON, no additional text.
\end{tcolorbox}
\noindent \begin{minipage}{\columnwidth}
\captionof{figure}{Judge prompt for RNR Metric}
\label{fig_prompts:judge_rnr}
\end{minipage}

\begin{tcolorbox}[enhanced jigsaw,breakable,pad at break*=1mm,
  colback=red!5!white,colframe=red!75!black,title=User-Proxy System Prompt]
You are simulating a real human user for the \textsc{MirrorBench} evaluation harness. Respond with the next USER turn only. Do not write assistant messages, notes, or any other analysis. Your utterance should be like a real user and the context should be based on the following information provided. \\

Task description: \{task\_description\} \\

Match the length, tone, and specificity of real user utterances. If you are unsure, respond naturally based on the assistant's previous messages like how a real human would. Note that your response MUST not contain anything other than the USER utterance. Do not include any prefixes like 'User:' or 'Human:' as well. Just the raw message content.
\end{tcolorbox}
\noindent \begin{minipage}{\columnwidth}
\captionof{figure}{System prompt for User-Proxy}
\label{fig_prompts:prompt_user_proxy}
\end{minipage}

\begin{tcolorbox}[enhanced jigsaw,breakable,pad at break*=1mm,
  colback=red!5!white,colframe=red!75!black,title=Assistant System Prompt]
You are the assistant in a \textsc{MirrorBench} replay. The user-proxy agent is attempting to reproduce the USER side of the real conversation provided below. But the user-proxy does not have access to the real conversation history. Instead, it only has access to the conversation summary. \\

You need to respond as the assistant. But we are providing you with the real conversation history as context, so you can respond consistently same as (or similar to) the original assistant in the real conversation (you may paraphrase lightly for safety). \\

If user-proxy deviates from the original USER turn or the original response would violate policy, reply helpfully using your own knowledge while remaining consistent with the persona demonstrated so far. Always follow ethical content policies. Paraphrase sensitive content
instead of quoting it verbatim, and refuse politely if a request is disallowed.  \\

Here is the real conversation for context (the USER turns are from the original conversation):  

\{real\_conversation\}  \\

Now, we will provide you with ongoing conversation with the user-proxy. Please respond as the assistant in this conversation.
\end{tcolorbox}
\noindent \begin{minipage}{\columnwidth}
\captionof{figure}{System prompt for Assistant}
\label{fig_prompts:prompt_assistant}
\end{minipage}

Figure \ref{fig_prompts:prompt_user_proxy} and \ref{fig_prompts:prompt_assistant} show system prompts for user-proxy and assistant respectively. Note that once filled, the system prompt would go into chat history as system message. Thus, we do not see the chat history-related placeholder in these system prompts.

\subsection{Evaluation Config, Manifest \& Report}
\label{app:config_manifest_report}

\begin{figure*}[h]
    \centering
    \includegraphics[width=0.7\textwidth, trim = 0cm 1cm 0cm 1cm, clip]{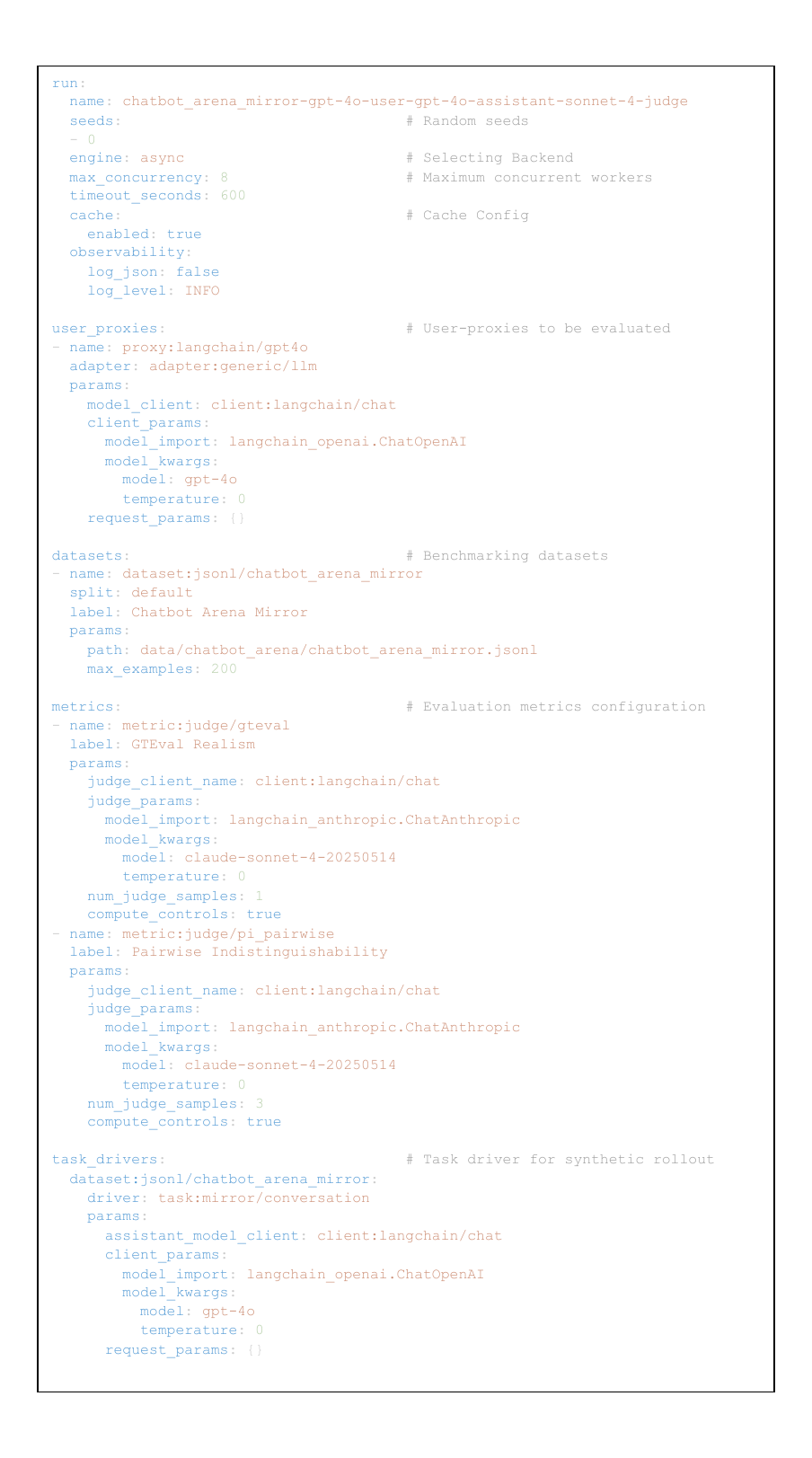}
    \caption{\textbf{Job configuration (abridged).} YAML that fully specifies a \textsc{MirrorBench} run: backend and concurrency, proxy/assistant clients, dataset binding, judge metrics (with calibration controls), and the mirror-conversation driver.}
    \label{fig:config_yaml}
\end{figure*}

Figure \ref{fig:config_yaml} shows the example job configuration. The configuration is declarative: the \texttt{run} block fixes execution semantics (backend, concurrency, timeouts), reproducibility (\texttt{seeds}), and infrastructure knobs (cache and logging). In our example, the async backend with \texttt{max\_concurrency=8}, while caching is enabled to deduplicate repeated LLM calls across units. The \texttt{user\_proxies} block registers a concrete proxy via a lightweight adapter and a model client (here, LangChain’s \texttt{ChatOpenAI} wrapper for GPT-4o), keeping proxy identity and invocation details separate from orchestration. The \texttt{datasets} block binds the ChatbotArena's \textsc{MirrorBench} corresponding split to a local JSONL path and caps evaluation at 200 examples for controlled runtime.

\begin{figure}[h]
    \centering
    \includegraphics[width=0.6\columnwidth, trim = 0cm 2.5cm 0cm 2cm, clip]{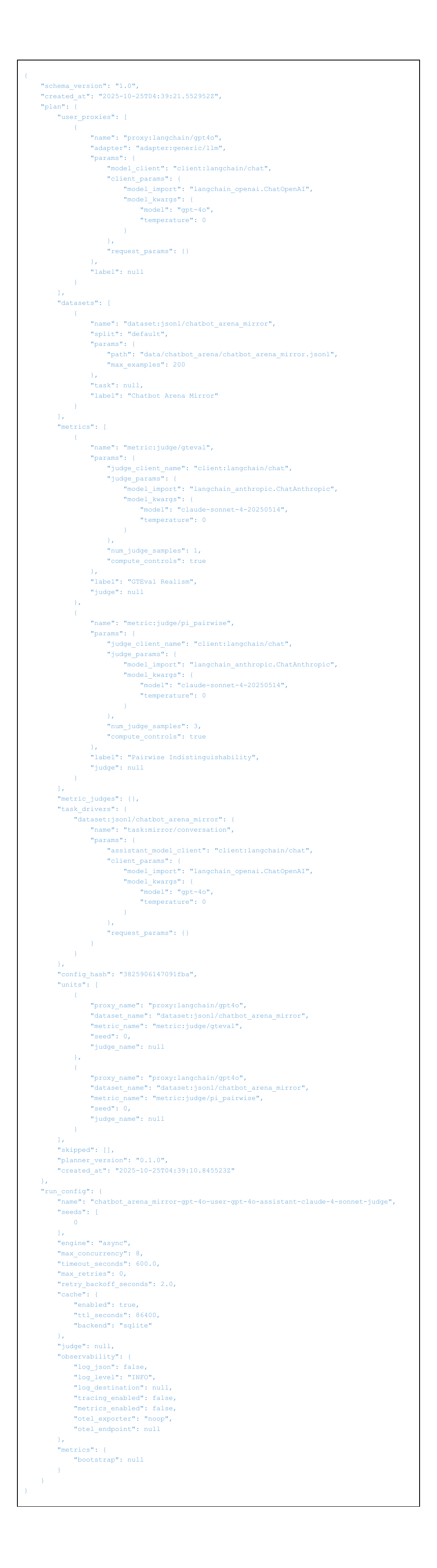}
    \caption{\textbf{Plan manifest (JSON, abridged).} Fully resolved, replayable spec listing the (proxy, dataset, metric, seed) units, driver params, versions, and \texttt{config\_hash}, enabling deterministic re-runs.}
    \label{fig:manifest_json}
\end{figure}

Metrics are declared under \texttt{metrics} with their own clients and prompts shown in \ref{app:system-prompts}, allowing the judge to differ from the proxy/assistant. We show two metrics, GTEval and Pairwise Indistinguishability, both routed to Claude-4-Sonnet at \texttt{temperature=0}. Each metric includes \texttt{num\_judge\_samples} (for repeated scoring) and \texttt{compute\_controls=true}, which triggers the HH/PP control comparisons used for calibration and bias checks. Finally, the task drivers section binds the dataset to the Mirror Conversation Driver, which synthesizes multi-turn dialogues by alternating the user-proxy and assistant models; here, the assistant is also GPT-4o at \texttt{temperature=0}, ensuring low-variance responses for per-proxy comparability. Together, these blocks yield a self-contained manifest that our planner expands into (proxy, dataset, metric, seed) units, enabling exact replay and variance-aware aggregation across runs. We illustrate an example of such manifest file in Figure \ref{fig:manifest_json}, which is a manifest produced using config shown in Figure \ref{fig:config_yaml}.

\begin{figure}[h]
    \centering
    \includegraphics[width=0.6\columnwidth, trim = 0cm 1.5cm 0cm 1.5cm, clip]{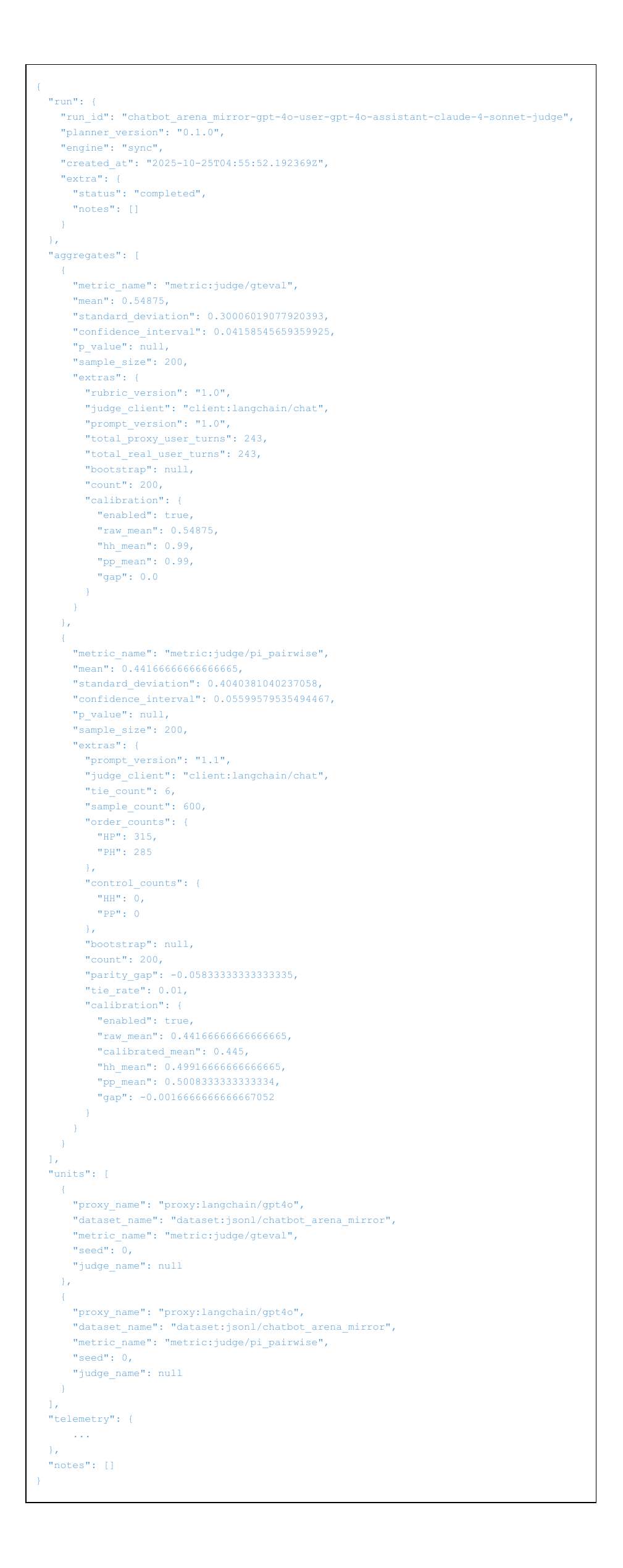}
    \caption{\textbf{Final run report (JSON, abridged).} Aggregate metric scores with 95\% CIs, calibration controls (HH/PP), unit grid, and run metadata for the completed evaluation.}
    \label{fig:report_json}
\end{figure}

Once the CLI command \texttt{mirrorbench run -c config.yaml} is invoked, the evaluation job starts and executes through all the units. Once the run is completed, another command \texttt{mirrorbench report json <run-name> --output report.json} can be executed to generate final evaluation report as shown in Figure \ref{fig:report_json}.

\clearpage

\end{document}